\RequirePackage{xcolor}
\documentclass[10pt,journal,twoside,web]{ieeecolor}
\usepackage{generic}
\usepackage{cite}
\usepackage{amsmath,amssymb,amsfonts}
\usepackage{algorithmic}
\usepackage{graphicx}
\usepackage{textcomp}
\usepackage{color,soul}
\usepackage{cuted}
\usepackage{kotex}
\usepackage{booktabs}
\usepackage{bbm}
\usepackage{amssymb}
\usepackage{enumerate}
\usepackage{hyperref}
\usepackage{longtable}
\hypersetup{colorlinks,allcolors=black}

\newcommand{\model}{MedViLL\xspace}
\usepackage{bm,amsmath}
\usepackage{xspace}
\usepackage{graphicx}
\usepackage{color}
\usepackage{algorithm}
\usepackage{algorithmic}
\usepackage{colortbl}

\usepackage{url}            % simple URL typesetting
\usepackage{booktabs}       % professional-quality tables
\usepackage{amsfonts}       % blackboard math symbols
\usepackage{nicefrac}       % compact symbols for 1/2, etc.
\usepackage{microtype}      % microtypography
\usepackage{longtable,lscape}
\usepackage{array,multirow}

\usepackage{subcaption}
\usepackage{wrapfig}

\newcommand{\hide}[1]{}

\let\labelindent\relax
\usepackage{enumitem}

\newcommand{\Hb}{\mathbf{H}}

\newcommand{\lb}{\mathbf{l}}

\newcommand{\pb}{\mathbf{p}}

\newcommand{\sbb}{\mathbf{s}}

\newcommand{\vb}{\mathbf{v}}
\newcommand{\wb}{\mathbf{w}}

\usepackage{mathtools}

\def\BibTeX{{\rm B\kern-.05em{\sc i\kern-.025em b}\kern-.08em
    T\kern-.1667em\lower.7ex\hbox{E}\kern-.125emX}}
\begin{document}

\title{Multi-modal Understanding and Generation for Medical Images and Text via Vision-Language Pre-Training}

\author{Jong Hak Moon$^{*}$, Hyungyung Lee$^{*}$, Woncheol Shin, Young-Hak Kim, and Edward Choi
% \thanks{This work was supported by Samsung Research.}
\thanks{First authors*: Jong Hak Moon, and Hyungyung Lee are equally contributed. Corresponding author: Edward Choi.}
\thanks{Jong Hak Moon, Hyungyung Lee, Woncheol Shin, and Edward Choi are with EdLab, Graduate School of AI, KAIST, Daejeon, South Korea (e-mail: jhak.moon@kaist.ac.kr; ttumyche@kaist.ac.kr; swc1905@kaist.ac.kr; edwardchoi@kaist.ac.kr).}
\thanks{Young-Hak Kim is with Department of Cardiology, Asan Medical Center, University of Ulsan College of Medicine, Seoul, South Korea(e-mail: mdyhkim@amc.seoul.kr).}}
\maketitle

\begin{abstract}
Recently a number of studies demonstrated impressive performance on diverse vision-language multi-modal tasks such as image captioning and visual question answering by extending the BERT architecture with multi-modal pre-training objectives.
 In this work we explore a broad set of multi-modal representation learning tasks in the medical domain, specifically using radiology images and the unstructured report.
 We propose Medical Vision Language Learner (MedViLL), which adopts a BERT-based architecture combined with a novel multi-modal attention masking scheme to maximize generalization performance for both vision-language understanding tasks (diagnosis classification, medical image-report retrieval, medical visual question answering) and vision-language generation task (radiology report generation). By statistically and rigorously evaluating the proposed model on four downstream tasks with three radiographic image-report datasets (MIMIC-CXR, Open-I, and VQA-RAD), we empirically demonstrate the superior downstream task performance of MedViLL against various baselines, including task-specific architectures. The source code is publicly available at: https://github.com/SuperSupermoon/MedViLL
 
%  The source code is publicly available
% at:https://github.com/SuperSupermoon/MedViLL}
\end{abstract}


\section{Introduction}
\label{sec:introduction}
Vision-Language (VL) multi-modal research using radiographic images and associated free-text description (e.g., Chest X-rays and radiology report) is one of the most important and interesting works in the medical informatics \cite{biswal2020emixer},\cite{chapman2001comparison},\cite{kalpathy2010multimodal},\cite{litjens2017survey},\cite{liu2019clinically},\cite{schlegl2015predicting},\cite{wang2018tienet}. Although each VL modality provides different representations to the researcher, images and reports contain mutually helpful semantic information. Consequently, advances in VL multi-modal research can be beneficial in improving the quality of clinical care by providing automated support for a variety of tasks such as diagnosis classification  \cite{chapman2001comparison},\cite{kalpathy2010multimodal},\cite{wang2018tienet}, report generation \cite{liu2019clinically},\cite{wang2018tienet}. Owing to the high dimensionality, heterogeneity, and systemic biases, however, 
handling both image and clinical report to learn joint representation poses significant technical challenges.
\begin{figure}[ht!]
\captionsetup{font=footnotesize}
\includegraphics[width=\linewidth,]{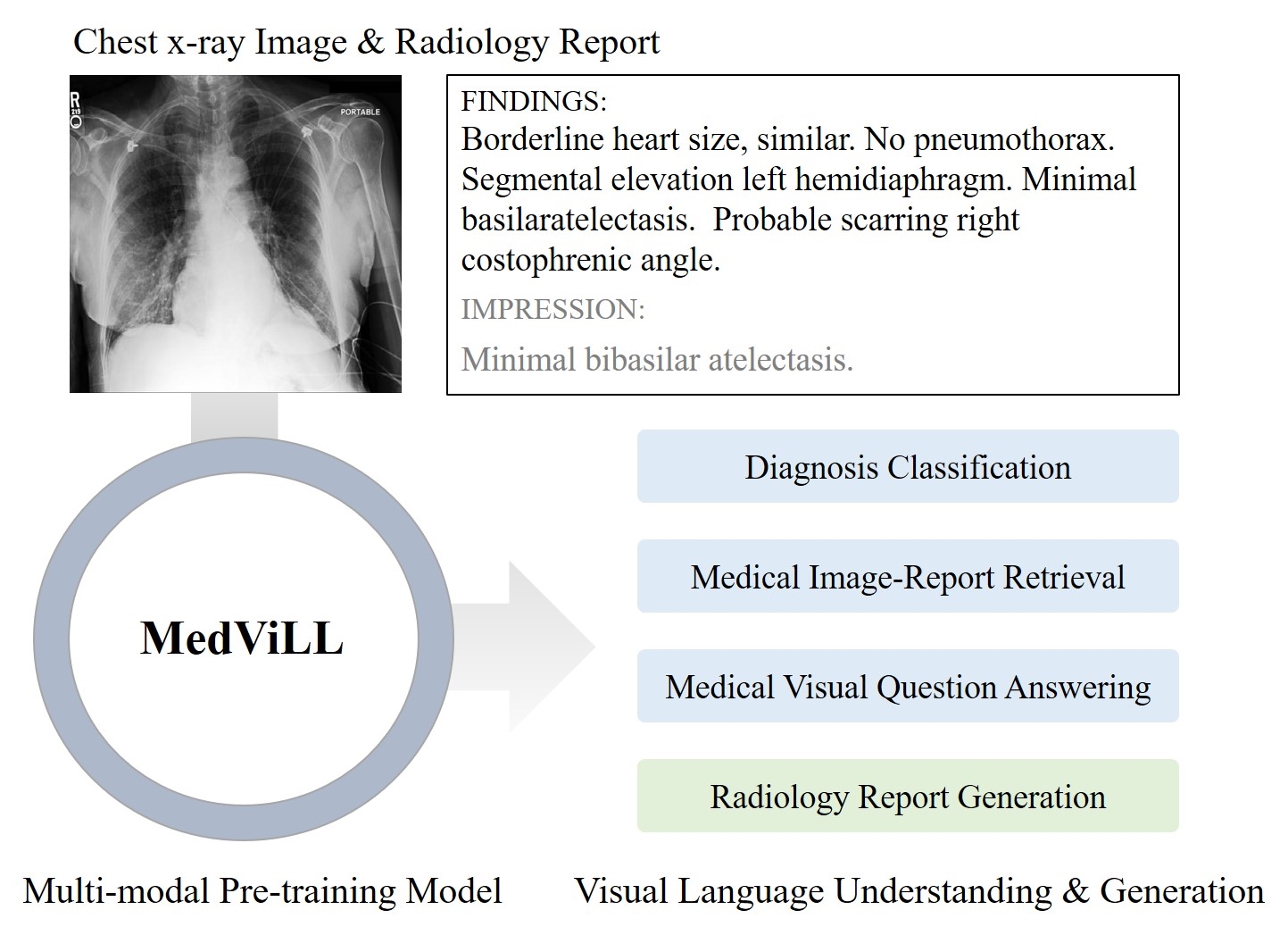}
\caption{\textbf{Overview of MedViLL.} During the pre-training, MedViLL learns joint representation, then fine-tuned for VLU and VLG tasks.}
\label{fig:fig1}
\end{figure}
The development of VL multi-modal learning has produced tremendous progress recently by extending the BERT-based architecture \cite{devlin2018bert} in deep learning area. BERT-based VL model is the typical pretrain-then-transfer approach that makes the model learn a representation of each modality by performing multiple pre-training tasks.
After pre-training the model, it is transferred to various vision language understanding (VLU) (e.g., visual question answering, text-conditioned image retrieval and vice versa) and vision language generation (VLG) (e.g., image captioning) downstream tasks by making only minor additions to the base architecture. 

However, despite significant improvements reported for a wide range of downstream tasks utilizing pre-trained models, most previous studies focused on either the VLU tasks or the VLG tasks \cite{huang2020pixel},\cite{li2019visualbert},\cite{lu2019vilbert},\cite{su2019vl}, \cite{tan2019lxmert}, suggesting the challenging nature of learning meaningful representations for both VLU and VLG at the same time. Some recent studies tried to tackle both tasks at the same time by proposing a hybrid model using the encoder and decoder of the Transformer \cite{sun2019learning},\cite{sun2019videobert},\cite{xia2021xgpt}, or with a unified BERT model by sharing knowledge using different types of the self-attention mask \cite{zhou2020unified}. These works demonstrated promising results even when a single BERT-based architecture was trained to aim at both tasks. 

While VL multi-modal pre-training has no doubt seen significant progress in recent years, it was mainly developed under the context of general domain (e.g., using MS-COCO). Vision and language, however, is one of the most frequently used information in the medical domain as well, often produced in the form of radiology images and corresponding free-text report. VL multi-modal pre-training therefore has great potential to be widely used in healthcare such improving diagnosis accuracy, automatically generating reports, or answering questions from physicians. Despite its huge potential, VL multi-modal pre-training in the medical domain has only recently received attention, where Li et al. \cite{li2020comparison} only demonstrated improved diagnosis accuracy of VL pre-trained models. In order to truly understand, however, whether a model has effectively learned both vision and text representation, it must be evaluated on diverse VLU and VLG tasks beyond simple diagnosis classification. This motivates us to investigate whether an integrated model is possible for a wide range of VLU and VLG tasks.

In this paper, we aim to develop a model that can learn multipurpose joint representations of vision and language in the medical domain (\hyperref[fig:fig1]{Fig.1}). The more immediate focus of our approach to enhance downstream tasks such as diagnosis and treatment delivery is case-based reasoning, discovering underlying patterns in data, and generating semantically accurate disease profiles.
%{Specifically, we use the Chest X-ray images and their corresponding reports, which are one of the most commonly taken radiology images, with large-scale datasets readily available online \cite{demner2016preparing}, \cite{johnson2019mimic}. However, these complex tasks in a unified framework are technically demanding because medical imaging and reports consist of multiple heterogeneous forms of information and the disparate tasks of VLU and VLG tasks. In order to encourage effectively multi-modal representation learning, we evaluate multiple self-attention methods, including a novel method of our own.}
The main contributions of this paper can be summarized as follows:
\begin{enumerate}
    \item We propose Medical Vision Language Learner (MedViLL), a multi-modal pre-training model for medical images and reports with a novel self-attention scheme.
    \item We demonstrate the effectiveness of our approach with detailed ablation study on extensive vision-language understanding and generation-based downstream tasks, including diagnosis classification, medical image-report retrieval, medical visual question answering, and radiology report generation.
    \item We demonstrate the generalization ability of our approach under the transfer learning setting using two separate Chest X-ray datasets, where we pre-train a model on one dataset and perform diverse downstream tasks on another.
\end{enumerate}

To the best of our knowledge, this is the first study that conducts both VLU and VLG tasks with a unified VL pre-training model in the medical domain. We expect that our pretrained VL model will enable more effective cross-task knowledge sharing, and reduce the development costs by eliminating the need for separate models for different tasks.

\section{Related Work}
\label{sec:relatedwork}
\subsection{Radiology Practices}
In radiology practice, physicians identify various clinical findings based on radiographic images and the patient's clinical history, then summarize these findings and overall impressions in a clinical report \cite{kahn2009toward}, \cite{schwartz2011improving}.
To accelerate the diagnostic process, \cite{sharif2020learning}, \cite{sharif2022deep} enhance perceptual quality of noisy radiographic images.
Diagnostic observations are described as positive, negative, or uncertain about the clinical findings, including the detailed location and severity of the findings. Such clinical reports are currently being used as a standard method to communicate in the clinical setting. A combination of vision and language data helps further improve the model performance in both image annotation and automatic report generation \cite{litjens2017survey}.

\subsection{VL Multimodal Researches in the Medical Domain}
Although various models have been gradually developed for language modeling \cite{alsentzer2019publicly},\cite{lee2020biobert},\cite{liu2021use}, \cite{smit2020chexbert}, CNN-RNN based models still dominate in VL multi-modal learning in the medical domain, and these models were mainly designed for a task-specific method of either VLU or VLG tasks. TieNet \cite{wang2018tienet} is a pioneering CNN-RNN model with image-report attention mechanism for VLU (e.g., diagnosis classification) and VLG (e.g., report generation) tasks by using ChestX-ray14 \cite{wang2017chestx} dataset. Liu et al. \cite{liu2019clinically} only focus on the VLG task to generate the radiology report utilizing a CNN-RNN-RNN architecture with a hierarchical generation strategy from the MIMIC-CXR \cite{johnson2019mimic} and Open-I dataset \cite{demner2016preparing}. Hsu et al. \cite{hsu2018unsupervised} focuses on a VLU task, specifically image-report retrieval in the MIMIC-CXR dataset, based on supervised and unsupervised methods. The most recent studies\cite{biswal2020emixer}, \cite{liu2021exploring}, \cite{wang2021self}, \cite{yang2021writing} focus on either VLU or VLG task. Li et al. \cite{li2020comparison} compares 4 different BERT-based pre-training models on a VLU task, specifically classifying thoracic findings in the MIMIC-CXR and Open-I dataset. With a focus on VLG task \cite{biswal2020emixer}, \cite{liu2021exploring}, \cite{wang2021self}, \cite{yang2021writing}, EMIXER \cite{biswal2020emixer} is a GAN-based approach that simultaneously generates a pair of X-ray images and corresponding reports based on diagnosis labels. Other recent VLG studies \cite{biswal2020emixer}, \cite{liu2021exploring}, \cite{wang2021self}, \cite{yang2021writing} focus on report generation that is as close to the ground truth as possible utilizing both the frontal and lateral view images to generate a single corresponding report. Liu et al. \cite{liu2021exploring} proposed a Transformer encoder-decoder based prior and posterior knowledge distilling approaches using 3 different modalities (Vision, Language, and Knowledge Graph). Wang et al. \cite{wang2021self} proposed a self-boosting framework with 3 different modules (CNN-based object detector as image encoder, Sentence-BERT as report encoder, and additional LSTM layers as decoder) using image and report based on the cooperation of the main tasks of generation and an auxiliary task of image-report matching. Yang et al. \cite{yang2021writing} proposed MedWriter that incorporates a hierarchical retrieval mechanism to automatically extract both report and sentence-level templates. In this paper, we focus on learning a joint representation of a single image (frontal view) and its corresponding report to perform both VLU and VLG tasks with fine-tuning.

\subsection{VL Multimodal Researches in the General Domain}
For better understanding of VL multimodality, many works have been proposed recently \cite{chen2019uniter}, \cite{huang2020pixel}, \cite{lu2019vilbert}, \cite{sun2019videobert}, \cite{tan2019lxmert}, \cite{zhou2020unified} in the general domain. Among numerous variants of VL pre-training setup, we focus on three components that are most relevant to our approach: input embedding stream, visual feature embedding, and downstream tasks.

\subsubsection{Input Embedding Stream}
Existing models can be divided into two groups based on their architecture as a single- \cite{chen2019uniter}, \cite{huang2020pixel}, \cite{li2019visualbert}, \cite{sun2019videobert}  or two-stream \cite{lu2019vilbert}, \cite{tan2019lxmert} with the marginal difference in downstream task performance \cite{bugliarello2020multimodal}, \cite{qi2020imagebert}. However, the two-stream architecture has a greater number of parameters, whereas the single stream architecture allows early interaction between two modalities by sharing parameters and processing stacks \cite{chen2019uniter}, \cite{huang2020pixel}, \cite{su2019vl}, \cite{zhou2020unified}. For architectural simplicity and time/space efficiency, we design our model with a single-stream architecture.

\subsubsection{Visual Feature Embedding}
For visual feature embedding, most of the recent works \cite{chen2019uniter}, \cite{li2019visualbert}, \cite{lu2019vilbert} are inspired by \cite{anderson2018bottom} utilizing pre-trained object detectors \cite{ren2016faster} to extract the region-based visual inputs. However, the representation capability of this approach is limited by the given categories of the object detection task, leading to information gaps for language understanding \cite{huang2020pixel}. In contrast to the region-based visual embedding, PixelBERT \cite{huang2020pixel} suggests CNN-based visual encoder with random pixel sampling to improve the robustness of visual feature learning and avoid over-fitting \cite{he2016deep}. Since there is no applicable off-the-shelf object detector model to extract region-based feature in the medical domain \cite{cheplygina2019not}, \cite{litjens2017survey}, \cite{raghu2019transfusion}, we adopt the CNN-based visual feature embedding.

\subsubsection{Downstream tasks}
VLU and VLG tasks are typical downstream tasks of the VL pre-trained model for tackling more complex tasks that combine vision with language. In this regard, a number of previous works \cite{chen2019uniter}, \cite{huang2020pixel}, \cite{lu2019vilbert}, \cite{tan2019lxmert} use BERT-based vision-language joint encoder to perform VLU tasks. On the other hand, VLG tasks typically require an encoder for embedding the vision features and a decoder that generates text \cite{sun2019videobert}. Unified VLP \cite{zhou2020unified} conducts these two disparate tasks (VLU and VLG) with a single BERT-based architecture by repeatedly alternating the mask type with a fixed ratio between bidirectional and sequence-to-sequence mask during pre-training. Inspired by this unified pre-training approach, we explore different types of masks and their effects on diverse VLU and VLG downstream tasks.

\begin{figure*}[ht]
\captionsetup{font=footnotesize}
\centering
\includegraphics[height=17cm,width=14.5cm]{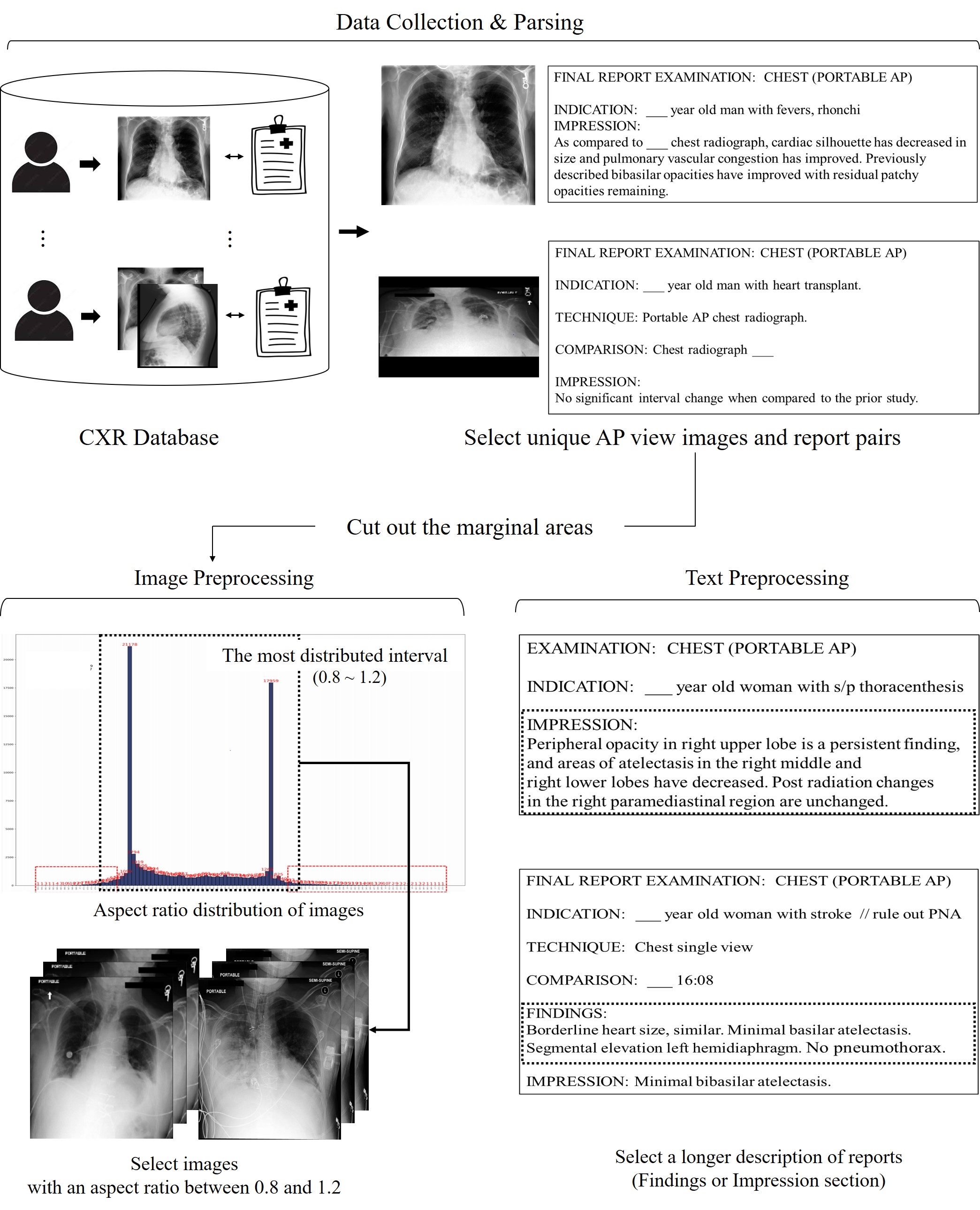}
\caption{\textbf{Data collection and parsing.} We pre-process the X-ray image and report data are as follows. First, for the X-ray image, we cut out the marginal space of the original image and resize all the images to 512 x 512, keeping the aspect ratio. Then for the report, we select a longer description (Findings or Impression section) which may contain detailed information associated with the X-ray imaging.}
\label{fig:fig2}
\end{figure*}

\section{Materials and Methods}
\label{sec:materialsandmethods}
\subsection{Dataset}
We used publicly available MIMIC-CXR \cite{johnson2019mimic} and Open-I \cite{demner2016preparing} datasets. MIMIC-CXR \cite{johnson2019mimic} contains 377,110 Chest X-ray images and corresponding free-text reports. Also, Open-I dataset contains 3,851 reports and 7,466 Chest X-ray images. Since the dataset contains frontal and lateral view images, it is required to distinguish between view positions \cite{chapman2001comparison}, \cite{liu2019clinically} to avoid miss-match findings between an image and a report pair. Therefore, given the dominance of the anteroposterior (AP) frontal view in ICU (Intensive Care Units) settings (e.g., 38.89\% of all studies containing at least one AP view image), we perform all experiments on unique 91,685 AP view image and associated report pairs following the official split of MIMIC-CXR (train 89,395, valid 759, test 1,531) and 3,547 image-report pairs from the official Open-I dataset. We use Open-I to test the generalization ability of the models, where all models are pre-trained on MIMIC-CXR, then fine-tuned for downstream tasks on a completely unseen Open-I dataset.
Our pre-processing procedure is illustrated in \hyperref[fig:fig2]{Fig. 2}.
% We pre-process the X-ray image and report data are as follows. First, for the X-ray image, we cut out the marginal space of the original image and resize all the images to 512 x 512, keeping the aspect ratio. Then for the report, we select a longer description (Findings or Impression section) which may contain detailed information associated with the X-ray imaging.

\begin{figure*}[htb!]
\centering
\captionsetup{font=footnotesize}
% [height=7cm, width=15cm]
\includegraphics[height=8.5cm,width=15cm]{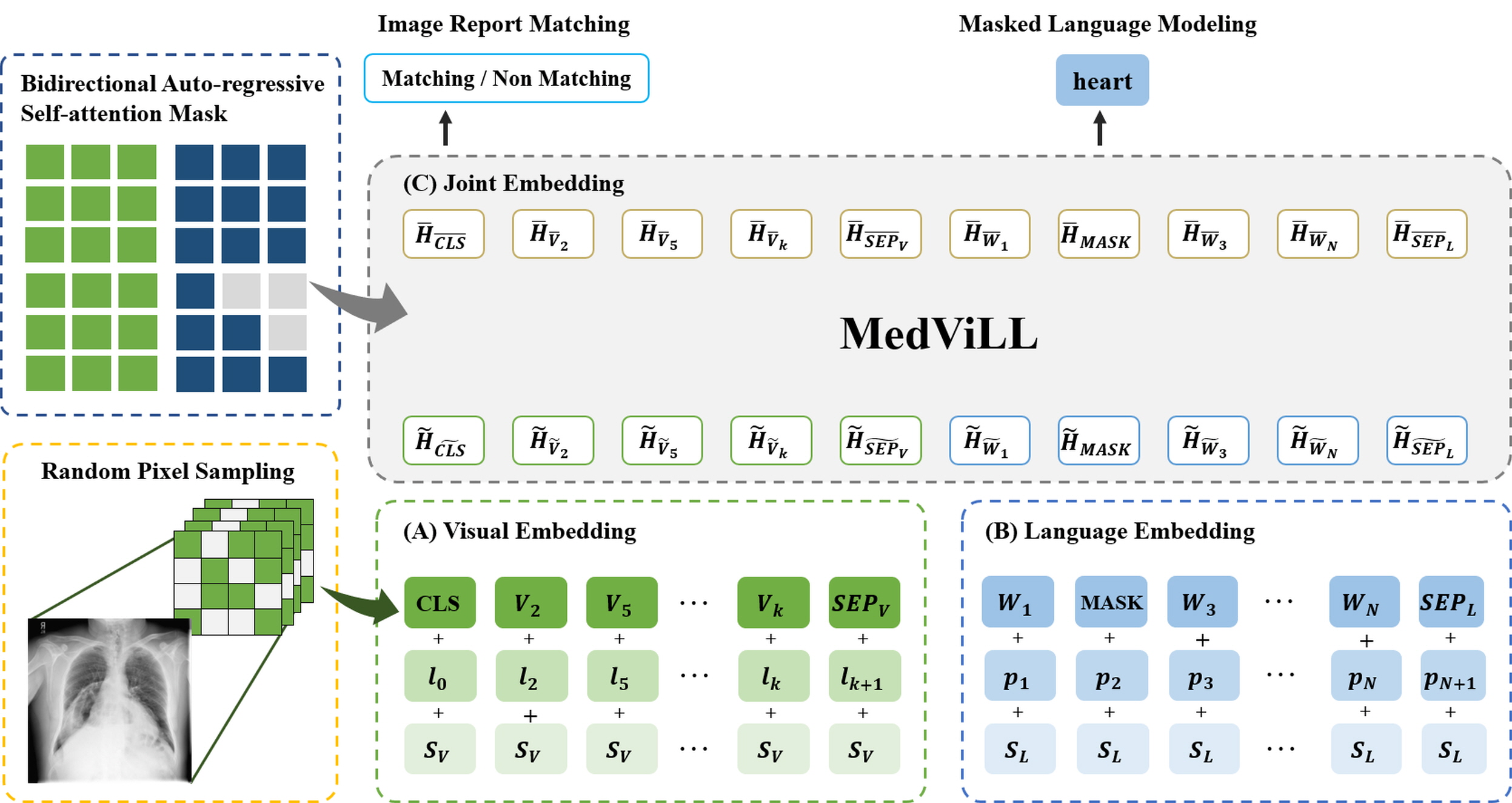}
\caption{\textbf{Architecture of the \model.} \model is a single stream BERT model for the cross-modal embedding. Chest X-ray images are randomly sampled from the last feature map of the CNN model as visual inputs. Also, each report is parsed with the BERT tokenizer to get language input. \model is pre-trained with masked language modeling and image report matching tasks, and flexibly applied to VLU and VLG downstream tasks.}

\label{fig:fig3}
\end{figure*}

\subsection{VL Pre-training Model}
Our proposed architecture MedViLL is a single BERT-based model that learn	unified contextualized vision-language representation. The overall architecture of MedViLL is illustrated in \hyperref[fig:fig3]{Fig.3}.

\noindent
\begin{table}[ht]
\footnotesize
\captionsetup{font=footnotesize}
\centering
\caption{Notation explanation appearing in this section}
\setlength{\tabcolsep}{3pt}
% \vspace{-3mm}
\begin{tabular}{cc}
\toprule
    Notation & Description \\
    \hline
    $v$ & Chest x-ray image\\
    $\vb$ & Visual feature\\
    $\lb$ & Visual location feature\\
    $\sbb_V$ & Visual semantic embedding\\
    $w$ & Clinical report \\ 
    $\wb$ & Language feature \\
    $\pb$ & Language position embedding\\
    $\sbb_L$ & Language semantic embedding \\
    $\tilde{\vb}$ & Final visual feature embedding \\
    $\tilde{\wb}$ & Final language feature embedding \\
    $\tilde{\Hb}$ & Joint embedding \\
    $\bar{\Hb}$ & Contextualized embedding \\
\bottomrule
\end{tabular}
\label{tab:term_explanation}
\end{table}%

\subsubsection{Visual Feature Embedding}
We use a CNN to extract visual features from the medical image. The visual features are obtained from the last convolution layer, then flattened along the spatial dimension. Further, we encode the absolute positions of visual input as additional information for explicitly injecting the same body position information in the x-ray images. Given a Chest x-ray image $v$, we denote the flattened visual feature obtained from the last CNN layer $\vb$, and the location feature $\lb$ as follows:
\begin{equation}
   \vb \,\text{=}\, \{\vb_1, \vb_2, \ldots, \vb_K\}, \enskip \vb_i \in \mathbb{R}^c
    \label{eq:cnn_output}
\end{equation}
\begin{equation}
    \lb \,\text{=}\, \{\lb_1, \lb_2, \ldots, \lb_K\}, \enskip \lb_i \in \mathbb{R}^c
    \label{eq:visual_location_feature}
\end{equation}
where $K$ indicates the number of visual features (\textit{i.e.}, height $\times$ width) and $c$ the hidden dimension size (\textit{i.e.}, channel size).
% we denote the flattened visual feature obtained from the last CNN layer as $\vb = \{\vb_1, \vb_2, \ldots, \vb_K\}, \enskip \vb_i \in \mathbb{R}^c$, and the location feature as $\lb = \{\lb_1, \lb_2, \ldots, \lb_K\}, \enskip \lb_i \in \mathbb{R}^c$ where $K$ indicates the number of visual features (\textit{i.e.} height $\times$ width) and $c$ the hidden dimension size (\textit{i.e.} channel size).
The final visual feature embeddings $\tilde{\vb} \,\text{=}\, \{\tilde{\vb}_1, \tilde{\vb}_2, \ldots, \tilde{\vb}_K\}$ are computed as follows:
\begin{equation}
    \tilde{\vb}_i \,\text{=}\, \vb_i + \lb_i + \sbb_V \label{eq:visual_feature}
\end{equation}
where $\sbb_V$ is a semantic embedding vector shared by all visual feature to differentiate themselves from language embeddings. The final visual features $\tilde{\vb}$ are fed into a fully-connected layer, to be projected into the same embedding space $\mathbb{R}^d$ has the language embeddings. During pre-training, we randomly sample a subset of the final visual features to avoid overfitting and enhance the semantic knowledge learning of visual input \cite{irvin2019chexpert}. We use $k$ to denote the number of sampled visual features, whereas $K$ denotes the number of all visual features.

\subsubsection{Language Feature Embedding}
{For language feature embedding, we follow BERT \cite{devlin2018bert} to encode the textual information.  A given clinical report $w$ is first split into a sequence of $N$ tokens (i.e. subwords) $\{w_1, ... ,w_N\}$ using the WordPiece tokenizer \cite{wu2016google}. The tokens are then converted to vector representations $\wb \,\text{=}\, \{\wb_1, \wb_2, \ldots, \wb_N\}, \enskip \wb_i \in \mathbb{R}^d$ via a lookup table, where $d$ is the embedding dimension size. We denote position embeddings as $\pb \,\text{=}\, \{\pb_1, \pb_2, \ldots, \pb_N\}, \enskip \pb_i \in \mathbb{R}^d$.  The final language feature embeddings $\tilde{\wb} \,\text{=}\, \{\tilde{\wb}_1, \tilde{\wb}_2, \ldots, \tilde{\wb}_N\}$ are obtained as follows:
\begin{equation}
    \tilde{\wb}_i \,\text{=}\, \wb_i + \pb_i + \sbb_L \label{eq:language_feature}
\end{equation}
where $\sbb_L$ is a semantic embedding vector shared by all language feature to differentiate themselves from visual embeddings.}

\subsubsection{Joint Embedding}
{After obtaining visual embedding $\tilde{\vb} \in \mathbb{R}^d$ and language embeddings $\tilde{\wb} \in \mathbb{R}^d$, we concatenate them to construct the input sequence to the joint embedding component (\hyperref[fig:fig3]{Fig.3} (C)). Using additional special tokens CLS and SEP, we define the input to the joint embedding block as $\tilde{\Hb} \,\text{=}\, \{\overline{CLS}, \tilde{\vb}_1, \ldots, \tilde{\vb}_K,\overline{SEP}_V, \tilde{\wb}_1, \ldots, \tilde{\wb}_N, \overline{SEP}_L\} \in \mathbb{R}^{d \times S}$ where $S \,\text{=}\, N + K + 3$. Note that $\overline{CLS}, \overline{SEP}_V$ and $\overline{SEP}_L$ are obtained by summing the special tokens with corresponding position and semantic embeddings as in \hyperref[fig:fig3]{Fig.3}. The contextualized embedding produced by the joint embedding block are denoted as $\bar{\Hb} \,\text{=}\, \{\overline{CLS}, \bar{\vb}_1, \ldots, \bar{\vb}_K, \overline{SEP}_V, \bar{\wb}_1, \ldots, \bar{\wb}_N, \overline{SEP}_L$\}.

\subsubsection{Pre-training Objectives}
{To pre-train MedViLL and align visual features with language features, we take the Masked Language Modeling (MLM) and Image Report Matching (IRM) tasks, which were used in various forms in previous work \cite{chen2019uniter}, \cite{huang2020pixel}, \cite{li2019visualbert}, \cite{su2019vl}. For the MLM task, we follow BERT to replace 15\% of the input text tokens $\{w_1, \ldots, w_N\}$ with the special MASK token, a random token, or the original token with a probability of 80\%, 10\% and 10\% respectively. The model is trained to recover these masked tokens based on the contextual observation of their surrounding language tokens and the visual tokens, by minimizing the following negative log-likelihood.
\begin{equation}
    L_{MLM}(\theta) \,\text{=}\, \mathbin{-}\mathbb{E}_{(v,w) \sim D} \Big[ \log P_\theta(w_m|v, w_{\setminus m}) \Big]
\end{equation}
where $\theta$ is the trainable parameters of \model.
A pair of images and its corresponding report $(v, w)$ is sampled from the training set $D$, where $w$ can be divided into the masked tokens $w_m$ and their complements $w_{\setminus m}$. $\mathbb{E}_{(v,w) \sim D}$ is the average for the training set $D$, and $P_\theta(w_m|v, w_{\setminus m})$ is the probability of $w_m$ given $v$ and $w_{\setminus m}$.
IRM task encourages the model to learn both visual and textual features by training the model to predict whether a given pair of image and report $(v, w)$ is a matching pair or not.
During pre-training, we randomly sample both matching image-report pairs and non-matching image-report pairs with 1:1 ratio from the dataset.  Note that, however, while selecting a matching pair is straightforward (X-ray images come with a corresponding report), sampling a non-matching pair is not, because two different report can be semantically the same (\textit{e.g.}, ``No findings.'', ``Nothing noticeable.'').
Therefore, when sampling for non-matching image-report pairs, we use diagnosis labels. Specifically, in our IRM task, a non-matching report is defined as the ones that are extracted different positive diagnosis labels than the matching report. The joint contextualized embedding $\overline{CLS}$ is used to classify whether the input image and report are a matching pair or not, with the following loss function,

\begin{align}
    L_{IRM}(\theta) \,\text{=}\, & \mathbin{-}\mathbb{E}_{(v,w) \sim D} \Big[ y \log P_\theta(v,w)\Big] \nonumber \\
    & -\mathbb{E}_{(v,w') \sim D} \Big[(1 - y) \log (1 - P_\theta(v,w') ) \Big]
\end{align}
where $(v, w)$ denotes a matching image-report pair, $(v, w')$ a non-matching pair, $y$ is the label (1 for matching, and 0 for unmatching), $\mathbb{E}_{(v,w) \sim D}$ is the average for the training set $D$, and $P_\theta(v,w)$ is the probability of the $(v, w)$ being paired.}

\begin{figure}[ht!]
\captionsetup{font=footnotesize}
\includegraphics[width=\linewidth]{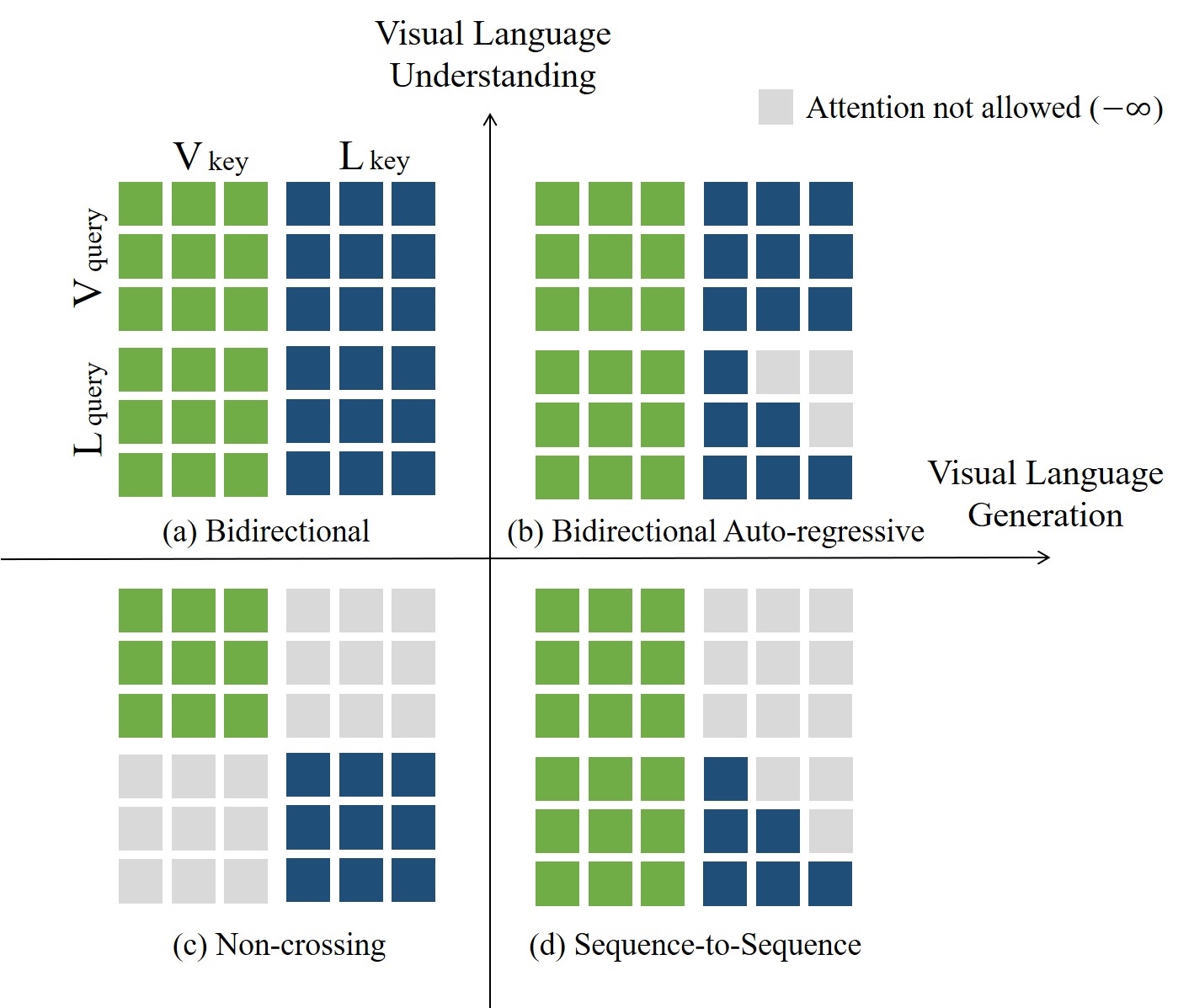}
\caption{\textbf{Self-attention mask schemes.} Four types of self-attention masks and the quadrant for the difference in performance in the downstream task of each attention mask. (a) Bidirectional, (b) Bidirectional Auto-regressive, (c) Non-crossing, (d) Sequence-to-Sequence self-attention masks.}

\label{fig:fig4}
\end{figure}

\subsection{Self-Attention Mask Schemes}
We explore several types of self-attention masks to encourage the model to learn universal multi-modal representations. Bi (Bidirectional) attention mask (\hyperref[fig:fig4]{Fig. 4} (a)) that allows all inputs to interact freely for unconstrained context learning between the visual-language modalities. S2S (Sequence-to-Sequence) causal attention mask (\hyperref[fig:fig4]{Fig. 4} (d)), on the other hand, allows restricted context learning; language features are only allowed to attend to previous words, while visual features are not allowed to attend to any language features, in order to prevent leaking information from the future. Bi \& S2S uses both Bi and S2S masks alternately during pre-training (in every mini-batch, use S2S with 75\% chance and Bi with 25\% chance) to perform both VLU and VLG downstream tasks. In this work, we propose a new self-attention mask, Bidirectional Auto-Regressive (BAR) (\hyperref[fig:fig4]{Fig. 4} (b)), to closes the gap between Bi and S2S while taking advantage of both. BAR allows image features to be mixed with language features during pre-training (as opposed to S2S mask), while preserving the causal nature of auto-regressive language generation.
% \begin{strip}
% \begin{align}
% \label{sa}
% SA = \frac{QK^{T}}{\sqrt{d_{k}}} = \frac{1}{\sqrt{d_{k}}}\begin{bmatrix}
% CLS_{q} \cdot CLS_{k}  & \cdots & CLS_{q} \cdot SEP_{Vk}&  CLS_{q} \cdot W_{1k}& \cdots &  CLS_{q} \cdot SEP_{Lk}\\
% \vdots & \ddots & \vdots& \vdots & \ddots & \vdots\\
% SEP_{Vq} \cdot CLS_{k} & \cdots & SEP_{Vq}\cdot SEP_{Vk}& SEP_{Vq} \cdot W_{1k} & \cdots & SEP_{Vq} \cdot SEP_{Lk}\\
% W_{1q} \cdot CLS_{k} & \cdots & W_{1q}\cdot SEP_{Vk}&W_{1q} \cdot W_{1k} & \cdots & W_{1q} \cdot SEP_{Lk}\\
% \vdots & \ddots& \vdots& \vdots & \ddots & \vdots\\
% SEP_{Lq} \cdot CLS_{k} &  \cdots & SEP_{Lq}\cdot SEP_{Vk}&SEP_{Lq} \cdot W_{Lk}& \cdots & SEP_{Lq} \cdot SEP_{Lk}\\
% \end{bmatrix}\in \mathbb{R}^{S\times{S}}
% \end{align}
% \end{strip}
The self-attention mask $M \in R^{S\times S}, S \,\text{=}\, N + K +3$ consists of 0s and negative infinities as below. 
\begin{equation}
    M_{jk} \,\text{=}\, 
        \begin{cases}
            0, \quad \enskip \text{(attention allowed)} \\
            \mathbin{-}\infty, \enskip \text{(attention not allowed)} \\
        \end{cases}
     \quad j, k \,\text{=}\, 1, ..., S.
\end{equation}
And a single attention head in the self-attention module can be formulated as follows:
\begin{equation}
    Attention\,\text{=}\,\operatorname{softmax}\left(SA+M\right) V,
    \quad SA \,\text{=}\,\frac{Q K^{T}}{\sqrt{d_{k}}}
\end{equation}
where $Q, K$, $V$, and $d_{k}$ indicate queries, keys, values, and dimension of queries and keys respectively \cite{vaswani2017attention}.
\begin{align}
\label{sa}
SA\,\text{=}\,\begin{bsmallmatrix}
CLS_{q} \cdot CLS_{k}  & \cdots &  CLS_{q} \cdot W_{1k}& \cdots &  CLS_{q} \cdot SEP_{Lk}\\
V_{1q} \cdot CLS_{k}  & \cdots &  V_{1q} \cdot W_{1k}& \cdots &  V_{1q} \cdot SEP_{Lk}\\
\vdots & \ddots & \vdots & \ddots & \vdots\\
SEP_{Vq} \cdot CLS_{k} & \cdots & SEP_{Vq} \cdot W_{1k} & \cdots & SEP_{Vq} \cdot SEP_{Lk}\\
W_{1q} \cdot CLS_{k} & \cdots &W_{1q} \cdot W_{1k} & \cdots & W_{1q} \cdot SEP_{Lk}\\
W_{2q} \cdot CLS_{k} & \cdots &W_{2q} \cdot W_{1k} & \cdots & W_{2q} \cdot SEP_{Lk}\\
\vdots & \ddots& \vdots & \ddots & \vdots\\
W_{Kq} \cdot CLS_{k} &  \cdots &W_{Kq} \cdot W_{Lk}& \cdots & W_{Kq} \cdot SEP_{Lk}\\
SEP_{Lq} \cdot CLS_{k} &  \cdots &SEP_{Lq} \cdot W_{Lk}& \cdots & SEP_{Lq} \cdot SEP_{Lk}\\
\end{bsmallmatrix}
\end{align}}

% \begin{align}
% \label{sa}
% \colorbox{yellow}{SA}=\begin{bsmallmatrix}
% CLS_{q} \cdot CLS_{k}  & \cdots &  CLS_{q} \cdot W_{1k}&  CLS_{q} \cdot W_{1k}& \cdots &  CLS_{q} \cdot SEP_{Lk}\\
% V_{1q} \cdot CLS_{k}  & \cdots &  V_{1q} \cdot W_{1k}&  V_{1q} \cdot W_{1k}& \cdots &  V_{1q} \cdot SEP_{Lk}\\
% \vdots & \ddots & \vdots& \vdots & \ddots & \vdots\\
% SEP_{Vq} \cdot CLS_{k} & \cdots & SEP_{Vq} \cdot W_{1k}& SEP_{Vq} \cdot W_{1k} & \cdots & SEP_{Vq} \cdot SEP_{Lk}\\
% W_{1q} \cdot CLS_{k} & \cdots &W_{1q} \cdot W_{1k}&W_{1q} \cdot W_{1k} & \cdots & W_{1q} \cdot SEP_{Lk}\\
% W_{2q} \cdot CLS_{k} & \cdots &W_{2q} \cdot W_{1k} &W_{2q} \cdot W_{1k} & \cdots & W_{2q} \cdot SEP_{Lk}\\
% \vdots & \ddots& \vdots & \vdots& \ddots & \vdots\\
% W_{Kq} \cdot CLS_{k} &  \cdots &W_{Kq} \cdot W_{Lk}&W_{Kq} \cdot W_{Lk}& \cdots & W_{Kq} \cdot SEP_{Lk}\\
% SEP_{Lq} \cdot CLS_{k} &  \cdots &SEP_{Lq} \cdot W_{Lk}&SEP_{Lq} \cdot W_{Lk}& \cdots & SEP_{Lq} \cdot SEP_{Lk}\\
% \end{bsmallmatrix}
% \end{align}}

% 0  & \cdots &  0 & \cdots &  0\\
% 0 & \cdots &  0 & \cdots &  0\\
% \vdots & \ddots & \vdots & \ddots & \vdots\\
% 0 & \cdots &  0 & \cdots &  0\\
% 0 & \cdots &  -\infty & \cdots &  -\infty\\
% 0 & \cdots &  0 & \cdots &  -\infty \\
% \vdots & \ddots & \vdots & \ddots & \vdots\\
% 0 & \cdots &  0 & \cdots & -\infty \\
% 0 & \cdots &  0&  \cdots & 0\\

where $q$ and $k$ indicate query and key vectors respectively. Since the self-attention matrix is computed from the query and key vectors of vision-language modalities according to the Eq. \hyperref[sa]{(9)}, the computed self-attention matrix can be divided into 4 subparts of queries and key combinations by modality type.
\begin{align}
SA_{q,k} & \,\text{=}\,
  SA_{CLS_{q}:SEP_{Vq}, CLS_{k}:SEP_{Vk}} \\ 
&+ SA_{CLS_{q}:SEP_{Vq}, W_{1k}:SEP_{Lk}}  \\
&+ SA_{W_{1q}:SEP_{Lq}, CLS_{k}:SEP_{Vk}}  \\
&+ SA_{W_{1q}:SEP_{Lq},W_{1k}:SEP_{Lk}}
\end{align}

where Eq. (10) is the attention of query and key from vision, Eq. (11) is an attention mask of query from the vision and key from language, Eq. (12) is an attention mask of query from the language and key from vision, and Eq. (13) is an attention mask of query and key from language features. We combine the attention mask matrix M for the subparts of $SA$ because adding negative infinity to the calculated attention value will result in zero in the softmax operation.

\begin{equation}
       BAR_M \,\text{=}\, \small{\begin{bmatrix}
    0  & \cdots &  0 & \cdots &  0\\
    0 & \cdots &  0 & \cdots &  0\\
    \vdots & \ddots & \vdots & \ddots & \vdots\\
    0 & \cdots &  0 & \cdots &  0\\
    0 & \cdots &  -\infty & \cdots &  -\infty\\
    0 & \cdots &  0 & \cdots &  -\infty \\
    \vdots & \ddots & \vdots & \ddots & \vdots\\
    0 & \cdots &  0 & \cdots & -\infty \\
    0 & \cdots &  0&  \cdots & 0\\
    \end{bmatrix}\in \mathbb{R}^{S\times{S}}}
\end{equation}

Therefore, BAR attention mask (Eq. (14)) allows the attention calculations of all possible combinations except for the Eq. (13). Intuitively, this self-attention mask scheme applies auto-regressive attention masks to language modality to enhance joint embedding between vision and language modalities and perform well in both generation and understanding tasks. We implemented four different models each using different types of self-attention masks during pre-training; Bi, S2S, BAR and Bi \& S2S.
In addition, we also experiment with Non-crossing attention mask (\hyperref[fig:fig4]{Fig. 4} (c)) as a baseline to investigate the impact of multi-modal representation learning. As non-crossing attention mask restricts the interaction between two modalities, we add one additional CLS token at the beginning of the language features, so that both $\overline{CLS}_V$ and $\overline{CLS}_L$ can be used for the IRM pre-training task. Three types of self-attention mask matrices (Bi, S2S and Non-crossing) are described in the appendix.

\section{RESULTS AND DISCUSSION}
\label{sec:resultsanddiscussion}

\begin{figure*}[ht!]
\captionsetup{font=footnotesize}
\centering
\includegraphics[height=7cm, width=15cm]{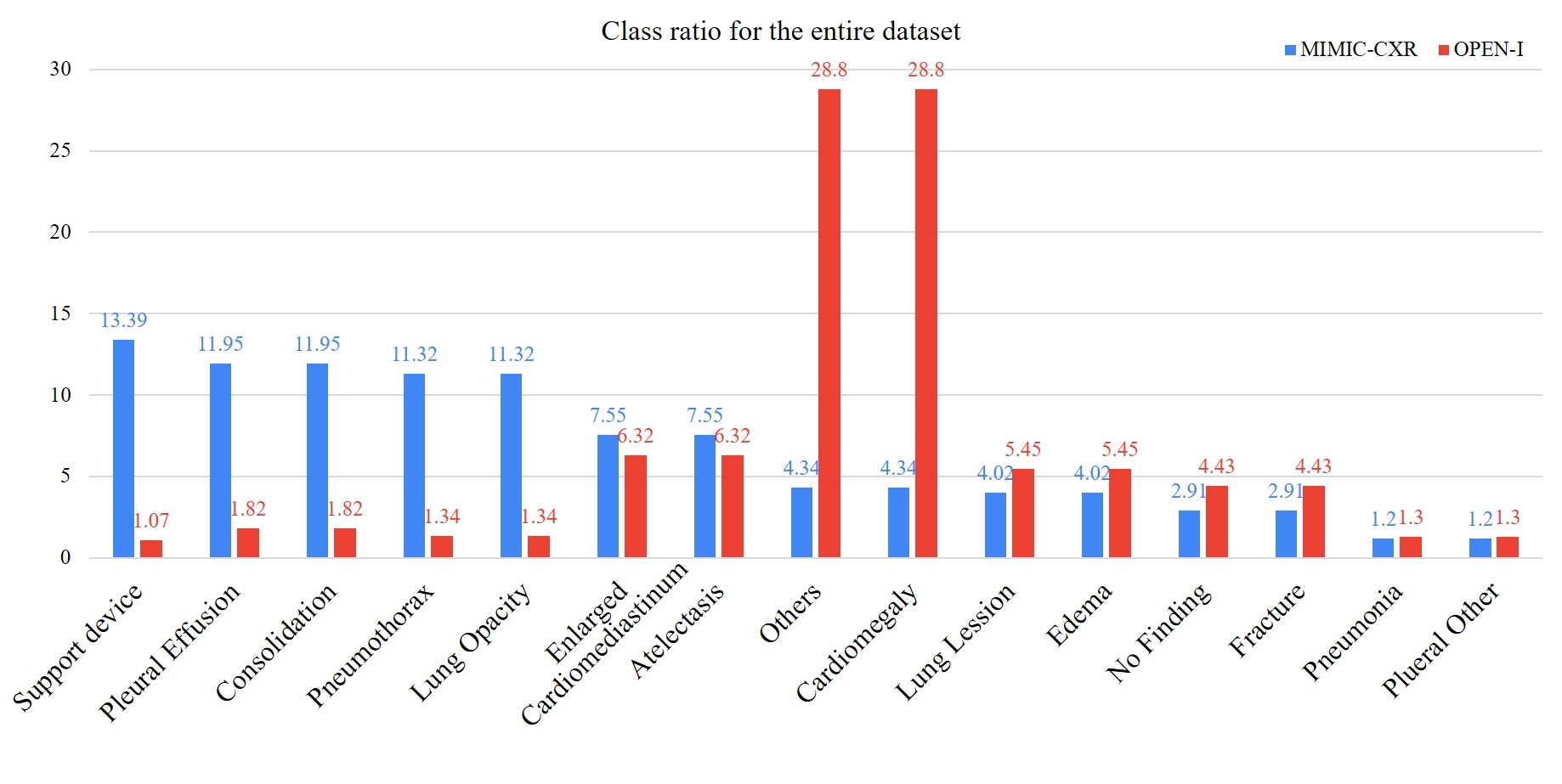}
\caption{\textbf{Dataset Analysis.} We compare the distribution of diagnosis labels over the entire dataset. Due to the different scales of the two datasets, each label was represented as a percentage over the entire dataset.}
\label{fig:fig5}
\end{figure*}

% \begin{table*}[h]
% \footnotesize
% \captionsetup{font=footnotesize}
% \centering
% \caption{Model AUROC and F1 scores for the diagnosis classification task on MIMIC-CXR and Open-I. \hl{Inference time(ms) on MIMIC-CXR: MedViLL(12.5), Bi\&S2S (13), Bi (13), S2S (13), Non-crossing (12.5), Fine-tuning Only (15.5), CNN \& Transformer (10.5).}}
% \setlength{\tabcolsep}{3pt}
% % \vspace{-3mm}
% \begin{tabular}{ccccccccc}
%     \toprule
%     Dataset & Metrics & \model & Bi\&S2S & Bi & S2S & Non-crossing & Fine-tuning Only & CNN \& Transformer   \\\hline
%     & avg AUROC & 0.982 (0.00) &  0.981 (0.00) & \textbf{ 0.986} (0.00) & 0.983 (0.00) &  0.981 (0.00) &  0.971 (0.00) & 0.832 (0.00) \\
%     MIMIC-CXR & avg F1 & 0.841 (0.01) & 0.848 (0.01) & \textbf{0.853} (0.01) & 0.847 (0.01) & 0.826 (0.00) &  0.809 (0.01) & 0.493 (0.01) \\
%     & \hl{p-value (avg AUROC)} & - & 0.005 & 1.97E-15 & 0.003 & 0.254 & 1.70E-36 & 3.41E-102 \\
%     & \hl{p-value (avg F1)} & - & 9.59E-28 & 7.85E-42 & 1.62E-26 & 2.02E-43 & 4.90E-63 & 2.70E-122 \\
%     \midrule
%     & avg AUROC	& \textbf{0.894 (0.00)} & 0.829 (0.00) & 0.759 (0.01) & 0.721 (0.01) & 0.591 (0.00) & 0.724 (0.00) & 0.710 (0.01) \\
%     Open-I & avg F1 & \textbf{0.408 (0.01)} & 0.302 (0.01) & 0.297 (0.01) & 0.257 (0.01) & 0.187 (0.00) & 0.301(0.00) & 0.246 (0.01) \\
%     & \hl{p-value (avg AUROC)} & - & 6.94E-83 & 4.00E-98 & 1.23E-101 & 4.66E-122& 1.41E-103 & 1.35E-109\\
%     & \hl{p-value (avg F1)} & - & 1.05E-93 & 7.04E-95 & 7.17E-101 & 2.08E-110 & 6.49E-94 & 2.69E-104\\
%     \bottomrule
% \end{tabular}
% \label{tab:classification}
% \end{table*}%

\begin{table*}[ht]
\footnotesize
\captionsetup{font=footnotesize}
\centering
\caption{Model AUROC and F1 scores for the diagnosis classification task on MIMIC-CXR and Open-I. Inference time(ms) on MIMIC-CXR: MedViLL(12.5), Bi\&S2S (13), Bi (13), S2S (13), Non-crossing (12.5), Fine-tuning Only (15.5), CNN \& Transformer (10.5).}
\setlength{\tabcolsep}{3pt}
% \vspace{-3mm}
\begin{tabular}{ccccccccc}
    \toprule
    Dataset & Metrics & \model & Bi\&S2S & Bi & S2S & Non-crossing & Fine-tuning Only & CNN \& Transformer   \\\hline
    & avg AUROC & 0.980 (0.00) &  0.979 (0.00) & \textbf{ 0.984 (0.00)} & 0.982 (0.00) &  0.980 (0.00) &  0.969 (0.00) & 0.831 (0.00) \\
    MIMIC-CXR & avg F1 & 0.839 (0.00) & 0.846 (0.00) & \textbf{0.852 (0.00)} & 0.846 (0.00) & 0.824 (0.00) &  0.807 (0.00) & 0.491 (0.00) \\
    & p-value (avg AUROC) & - & 0.005 & 1.97E-15 & 0.003 & 0.254 & 1.70E-36 & 3.41E-102 \\
    & p-value (avg F1) & - & 9.59E-28 & 7.85E-42 & 1.62E-26 & 2.02E-43 & 4.90E-63 & 2.70E-122 \\
    \midrule

    & avg AUROC	& \textbf{0.892 (0.00)} & 0.827 (0.00) & 0.758 (0.00) & 0.720 (0.00) & 0.589 (0.00) & 0.723 (0.00) & 0.709 (0.00) \\
    Open-I & avg F1 & \textbf{0.407 (0.01)} & 0.301 (0.01) & 0.295 (0.01) & 0.256 (0.01) & 0.185 (0.00) & 0.300(0.00) & 0.245 (0.01) \\
    & p-value (avg AUROC) & - & 6.94E-83 & 4.00E-98 & 1.23E-101 & 4.66E-122& 1.41E-103 & 1.35E-109\\
    & p-value (avg F1) & - & 1.05E-93 & 7.04E-95 & 7.17E-101 & 2.08E-110 & 6.49E-94 & 2.69E-104\\
    \bottomrule
\end{tabular}
\label{tab:classification}
\end{table*}%

\begin{table*}[ht]
\footnotesize
\captionsetup{font=footnotesize}
\centering
\caption{Medical Image-Report Retrieval performance on MIMIC-CXR and Open-I. Inference time(ms) of Report-to-Image and Image-to-Report on MIMIC-CXR: MedViLL(7.6, 7.8), Bi\&S2S (8.2, 7.8), Bi (7.6, 7.6), S2S (7.6, 7.7), Non-crossing (7.7, 7.8), Fine-tuning Only (7.8, 7.6), CNN \& Transformer (5.3, 5.3).}
\setlength{\tabcolsep}{3pt}
% \vspace{-3mm}
\begin{tabular}{cccccccccccc}
\toprule
    \multirow{2}{*}[-4pt]{Task} &
    \multirow{2}{*}[-4pt]{Models} &
    \multicolumn{5}{c}{MIMIC-CXR} &
    \multicolumn{5}{c}{OpenI} \\
    \cmidrule(lr){3-7}
    \cmidrule(lr){8-12}
    & & MRR & H@5 & R@5 & P@5 & p-value & MRR & H@5 & R@5 & P@5 & p-value\\
    \midrule
    & \model & 56.5(0.01) &77.0(0.01)&47.4(0.01) &19.9(0.00) & - & 51.3(0.01) &73.0(0.01) &12.9(0.00) &31.7(0.00) & - \\
    & Bi\&S2S & 55.5(0.01) &76.7(0.01) &46.7(0.01) &19.7(0.00) & 1.20E-05 & 46.4(0.01) &68.1(0.01) &10.5(0.00) &28.8(0.01) & 3.71E-27\\
    & Bi & 58.0(0.01) &78.2(0.01) &48.2(0.01) &20.2(0.00) & 1.60E-10 & \textbf{51.4(0.01)} &\textbf{74.8(0.01)} &\textbf{13.3(0.00)} &	32.0(0.01) & 0.843\\
    Report-to-Image & S2S & \textbf{58.8(0.01)} &\textbf{79.1(0.01}) &\textbf{48.9(0.01)} &\textbf{20.3(0.00)} & 1.89E-18 & 48.6(0.01) &67.2(0.01) &10.3(0.01) &\textbf{32.9(0.01)} & 2.28E-14\\
    & Non-crossing & 54.7(0.01) &77.0(0.01) &	47.2(0.01) &19.5(0.00) & 4.07E-12 & 48.6(0.01) &68.4(0.01) &11.2(0.00) &31.1(0.01) & 3.88E-18\\
    & Fine-tuning Only & 41.8(0.01) &61.6(0.01) &35.8(0.01) &15.8(0.00) & 3.14E-53 & 36.9(0.01) &54.4(0.01) &5.4(0.00) &20.7(0.01) & 1.72E-53\\
    & CNN \& Transformer & 11.4(0.01) &15.2(0.02) &5.1(0.00) &	3.6(0.01) & 9.43E-71 & 36.2(0.04) &56.6(0.04) &5.0(0.00) &	21.4(0.04) & 4.94E-19\\
    \midrule
    
    & \model & 55.8	(0.01) &75.5(0.01) &47.1(0.01) &19.7(0.00) & - & \textbf{50.4(0.01)} &63.8(0.01) &12.9(0.00) &35.5(0.01) & - \\
    & Bi\&S2S & 54.5(0.01) &75.5(0.01) &47.8(0.01) &19.9(0.00) & 6.32E-08 & 45.8(0.01) &54.0(0.01) &10.1(0.00) &35.8(0.00) & 8.55E-29 \\
    & Bi & 56.7(0.01) &76.3(0.01) &47.6(0.01) &20.2(0.00) & 0.0002 & 48.5(0.01) &\textbf{65.8(0.01)} &\textbf{13.7(0.00)} &32.3(0.01) & 3.17E-12 \\
    Image-to-Report & S2S & \textbf{57.9(0.01)} &\textbf{78.5(0.01)} &\textbf{49.7(0.01)} &\textbf{20.7(0.00)} & 2.72E-13 & 45.4(0.01) &53.6(0.01) &8.9(0.00) &\textbf{36.9(0.00)} & 6.84E-31 \\
    & Non-crossing & 54.6(0.01) &75.7(0.01) &47.6(0.01) &20.0(0.00) & 3.84E-07 & 42.6(0.01) &61.2(0.01) &11.0(0.00) &28.0(0.01) & 1.15E-40 \\
    & Fine-tuning Only & 41.4(0.01) &60.8(0.01) &36.3(0.01) &15.7(0.00) & 5.56E-56 & 45.2(0.00) &49.7(0.01) &5.1(0.00) &35.0(0.00) & 2.64E-29 \\
    & CNN \& Transformer & 12.0(0.02) &15.3(0.02) &5.1(0.00) &4.0(0.01) & 1.09E-52 & 37.9(0.06) &54.0(0.06) &5.0(0.00) &23.0(0.06) & 1.16E-12 \\
    
\bottomrule
\end{tabular}
\label{tab:retrieval}
\end{table*}%

\begin{table*}[ht]
\footnotesize
\captionsetup{font=footnotesize}
\centering
\caption{Model accuracy on the VQA-RAD dataset. O.E. stands for Open-ended question and C.E. stands for close-ended question. For MEVF \cite{nguyen2019overcoming}, we used the reported results from the original paper. Inference time(ms) on MIMIC-CXR: MedViLL(19.46), Bi\&S2S(19.52), Bi(19.43), S2S(19.51), Non-crossing(19.58), Fine-tuning Only(19.61), CNN \& Transformer(17.42).}
\setlength{\tabcolsep}{3pt}
% \vspace{-3mm}

\begin{tabular}{cccccccccc}
\toprule
    \multirow{2}{*}[-4pt]{Models} &
    \multicolumn{4}{c}{ALL} &
    \multicolumn{4}{c}{CHEST} \\
    \cmidrule(lr){2-5}
    \cmidrule(lr){6-9}
             & O.E. & C.E. & p-value of O.E. & p-value of C.E. & O.E. & C.E. & p-value of O.E. & p-value of C.E.\\
    \midrule
    MedViLL                         & \textbf{0.595(0.032)} &0.777(0.071) & - & - &  \textbf{0.587(0.033)} &\textbf{0.782(0.123)} & -& -\\
    Bi\&S2S                         & 0.541(0.038) &0.76(0.027) & 2.93E-07 & 0.224& 0.566(0.074) &0.766(0.035) &0.164&0.519 \\
    Bi                              & 0.58(0.038) &\textbf{0.784(0.03)} & 0.124 & 0.643 & 0.562(0.04) &0.767(0.035) & 0.013 & 0.549\\
    S2S                             & 0.505(0.042) &0.73(0.025) & 1.81E-12 & 0.002 & 0.517(0.07) &0.723(0.048) & 1.57E-05&0.021\\
    Non-crossing                    & 0.531(0.015) &0.734(0.017) & 5.58E-12 & 0.003 & 0.474(0.083) &0.732(0.03) & 3.94E-08 & 0.043\\
    Fine-tuning Only                & 0.232(0.019) &0.649(0.026) & 2.98E-43 & 5.38E-11 & 0.124(0.014) &0.606(0.035) & 1.08E-42  & 1.50E-08\\
    CNN \& Transformer               & 0.24(0.029) &0.667(0.015) & 7.66E-46 & 2.70E-9 & 0.124(0.067) &0.523(0.033) & 7.60E-32& 1.62E-12\\
    MEVF\cite{nguyen2019overcoming} & 0.407 & 0.741 & -  & - & -  &- &- & -\\
\bottomrule
\end{tabular}
%}
\label{tab:vqa}
\end{table*}%

\begin{table*}[ht]
\footnotesize
\captionsetup{font=footnotesize}
\centering
\caption{Report generation performance in terms of Perplexity and Label Accuracy, Precision Recall and F1 and BLEU4. Inference time(ms) on MIMIC-CXR: MedViLL(32.81), Bi\&S2S(33.55), Bi(32.54), S2S(33.04), Non-crossing(32.71), Fine-tuning Only(32.92).}
\setlength{\tabcolsep}{3pt}

% \vspace{-3mm}
\begin{tabular}{ccccccccccc}
\toprule
     Dataset & Models & Perplexity ($\downarrow$) & Accuracy ($\uparrow$) & Precision ($\uparrow$) & Recall ($\uparrow$) & F1 Score ($\uparrow$) & BLEU4 ($\uparrow$) & p-value\\
    \hline
    &MedViLL             & 4.185(0.022) &\textbf{0.841(0.003)} &\textbf{0.698(0.002)} &\textbf{0.559(0.004)} &\textbf{0.621(0.002)} &0.066(0.001) & - \\
    &Bi\&S2S         & 6.515(0.12) &0.786(0.007) &0.619(0.003) &0.435(0.009) &0.511(0.006) &0.066(0.001) & 4.17E-43 \\
    &Bi                  & 849.67(5.225) &0.637(0.004) &0.283(0.007) &0.07(0.024) &0.11(0.032) &0.015(0.004) & 1.11E-36 \\
    MIMIC &S2S             & 4.258(0.069) &0.797(0.007) &0.662(0.004) &0.448(0.01) &0.534(0.007) &0.043(0.001) & 2.32E-38\\
    &Non-crossing         & 718.122(9.484) &0.634(0.005) &0.277(0.013) &0.076(0.004) &0.12(0.005) &0.007(0.001) & 2.30E-75 \\
    &Fine-tuning Only    & 224.343(0.204) &0.664(0.003) &0.417(0.012) &0.305(0.006) &0.352(0.005) &0.009(0.004) & 4.14E-65 \\
    &TieNet               & \textbf{4.132(0.033)} & 0.687(0.003) & 0.487(0.003) & 0.380(0.006) & 0.426(0.006) & \textbf{0.123(0.002)} & 7.17E-54 \\
    
    \midrule
    &MedViLL             & 5.637(0.259) &0.734(0.001) &0.512(0.002) &0.594(0.001) &0.55(0.001) &0.049(0.001) & - \\
    &Bi\&S2S         & 15.97(1.071) &0.712(0.003) &0.497(0.003) &0.369(0.006) &0.423(0.004) &0.024(0.01) & 4.02E-52 \\
    &Bi                  & 787.66(55.492) &0.686(0.004) &0.356(0.025) &0.103(0.006) &0.16(0.008) &0.015(0.004) & 2.14E-52 \\
    Open-I &S2S             & \textbf{4.732(0.537)} &\textbf{0.736(0.003)} &\textbf{0.517(0.002)} &0.538(0.004) &0.527(0.002) &0.043(0.002) & 1.76E-44 \\
    &Non-crossing         & 217.27(12.139) &0.693(0.003) &0.337(0.025) &0.085(0.005) &0.135(0.007) &0.002(0.001) & 1.46E-57 \\
    &Fine-tuning Only    & 292.60(19.858) &0.684(0.003) &0.291(0.023) &0.073(0.035) &0.112(0.047) &0.006(0.002) & 7.02E-30 \\
    &TieNet               & 7.901(0.483) & 0.732(0.007) & \textbf{0.517(0.013)} & \textbf{0.610(0.017)} & \textbf{0.553(0.013)} & \textbf{0.189(0.005)} & 0.2181\\
\bottomrule
\end{tabular}
\label{tab:generation}
\end{table*}%

\subsection{Dataset Analysis}
Although both MIMIC-CXR and Open-I consist of chest X-ray images and report pairs, the two datasets could have different characteristics since they were collected from separate institutions. Specifically, the diagnostic information represented by the two X-ray image sets could be differently distributed. Therefore, to analyzes the difference in the distribution of diagnostic labels between two datasets, we compared positive labels acquired from the Chexpert labeler results. As seen in \hyperref[fig:fig5]{Fig. 5}, a mild imbalance was observed in MIMIC-CXR where the class ratios ranged from 13.39\% (support devices) to 1.2\% (pneumonia, and pleural other). On the other hand, a severe imbalance was observed in Open-I compared to MIMIC-CXR with the maximum class ratios of 28.8\% (Others, and cardiomegaly) and the minimum of 1.07\% (support devices). This shows that Open-I not only differs from MIMIC-CXR in terms of data volume, but also in terms of clinical properties. Therefore, we believe evaluating the MIMIC-CXR-pre-trained models on Open-I is an appropriate setup to test the generalization capability of the models. The distribution of diagnosis label is illustrated in \hyperref[fig:fig5]{Fig. 5}.

\subsection{Implementation details}We use ResNet-50 pre-trained on ImageNet as a visual feature extractor. The input image size is (512x512x3), and the last feature map (16x16x2048) of ResNet-50 is flattened by spatial dimensions and we randomly sample 180 visual features (180x2048) during pre-training, while we use all features (256x2048) for every downstream task. To embed text token, each sequence from reports is truncated or padded to 253 tokens in length by considering maximum embedding size. For the joint embedding, we adopt BERT-base architecture which comprised of 12 Transformer layers. Each layer contains 12 attention heads, 768 embedded hidden size and 0.1 drop-out probability. We adopt AdamW optimizer with learning rate $1e^{-5}$ settings for visual backbone and Transformer. All models were trained on 8 RTX-3090 GPU with the batch size of 128 and 50 epochs for the pre-training model.

\subsection{Task-specific Downstream Model Strategy}
\subsubsection{Diagnosis Classification}
{For a given image-report pair, we use the positive labels extracted from the report by the Chexpert labeler as the diagnosis labels. As a single pair could have multiple diagnosis labels up to the maximum of 14 (\textit{i.e.} multi-label classification), we use 14 linear heads on top of $\overline{CLS}$ and fine-tune the model using the binary cross-entropy loss. All models are evaluated with the micro average AUROC, and micro average F1 score.}

\subsubsection{Medical Image-Report Retrieval}
{There are two subtasks for medical image-report retrieval, where image-to-report (I2R) retrieval requires the model to retrieve the most relevant report from a large pool of reports given an image, and vice versa for report-to-image (R2I) retrieval. Given an image, any report that contains the same Chexpert diagnosis labels as the original matching report is considered a positive image-report pair, and a negative pair otherwise. The final multi-modal representation $\overline{CLS}$ is used as the input to a binary classifier to classify the given pair, which is trained by the binary cross-entropy loss. At inference, in each trial, a model is given 100 image-report pairs, and it must use the predicted scores to rank the positive pair as highest as possible. The evaluation metrics are Hit@K, Recall@K, Precision@K (K = 5), and mean reciprocal rank (MRR).}

\subsubsection{Medical Visual Question Answering}
{We perform VQA on the VQA-RAD dataset \cite{alizadehsani2019database}, which contains 3,515 question-answer pairs on 315 images (104 head CTs or MRIs, 107 Chest X-rays, and 104 abdominal CTs). As our models are pre-trained on Chest X-ray images, VQA-RAD provides a unique opportunity to study whether the pre-trained models would generalize well beyond the single image domain. Given a pair of an image and a free-text question, we use the final representation  $\overline{CLS}$ to predict a one-hot encoded answer (all possible answers are treated as a single token). The performance was evaluated with accuracy, but separately for the closed questions (i.e. short-form answers such as yes/no) and the open-ended questions (i.e. long-form answers), following the original VQA-RAD paper.}

\subsubsection{Radiology Report Generation}The fine-tuning process is same as the MLM pre-training task, except that we fix the self-attention mask to S2S for all models. At inference, reports can be generated by sequentially recovering the MASK tokens; given visual features followed by a single MASK token, the model can predict the first language token. Then we can replace the first MASK with the sampled token, and a new MASK token is appended. This process is repeated until the model predicts the SEP token as the stop sign. The performance is measured with three metrics: perplexity, clinical efficacy, and BLEU score. Clinical efficacy metrics is obtained by applying the Chexpert labeler on both the original matching report and the generated report. Based on the extracted labels, we can calculate accuracy, precision, recall, and F1 accordingly. Perplexity is used to evaluate the linguistic fluency of the model, while the clinical efficacy is used to evaluate if the model can capture the semantics of the given image. We also report 4-gram BLEU score to evaluate how similar the generated report is to the reference report.

\subsection{Downtream Task Result}MedViLL is compared to four pre-training models trained with different attention masks. We also include two more baselines: 1) Fine-tuning Only, which follows the same model architecture as MedViLL, but directly fine-tuned on each downstream task without any pre-training. 2) CNN \& Transformer, which uses the CNN module for encoding image only, and the Transformer module (same size as MedViLL) for encoding report only, and the outputs from each module are used for downstream tasks. CNN \& Transformer also does not use pre-training. For all tasks, we conduct 30 random multiple-bootstrap experiments and report the mean performance and its standard deviation. Also, we perform a statistical hypothesis test. Based on the average values of the various metrics obtained for each model, we conduct an independent t-test with a significance value of 0.05 to identify the significant pairwise difference of our method against multiple baseline models.  In summary, MedViLL achieved the best or second-best performance by analyzing statistical significance with various baseline models in the VLU and VLG tasks. In addition, MedViLL shows superior generalization ability by outperforming most of the models in out-of-domain evaluations.

\subsubsection{Diagnosis Classification}
All model performance is shown in \hyperref[tab:classification]{TABLE II}. For the Open-I images, we use Chexpert labeler on their MeSH annotations to extract the same set of diagnosis categories as the MIMIC-CXR dataset. Specifically, Bi and S2S outperform MedViLL with a statistically significant difference in both micro-averaged AUROC and F1 scores, indicating the null hypothesis can be rejected (t-test produced a p-value lower than 0.05). Also, although MedViLL (0.9805) achieves a higher score than Non-crossing (0.9801), it is not statistically meaningful with a p-value of 0.254 in micro-averaged AUROC. However, MedViLL outperforms all other baselines with a statistically meaningful difference in MIMIC-CXR. Moreover, MedViLL outperforms all models statistically significantly when transferred to Open-I. It is also noteworthy that Bi\&S2S, which is aimed to take advantage of both the bidirectional mask and the S2S mask demonstrates much better generalization capability compared to the two individual masks.

\subsubsection{Medical Image-Report Retrieval}
\hyperref[tab:retrieval]{TABLE III} shows the performance of Image-to-Report and Report-to-Image retrieval. We report the p-value of MRR for the performance of both tasks since MRR is a rank-aware evaluation metrics compared to other metrics.
We can observe that all pre-trained models significantly outperform naive baselines (Fine-tune Only and CNN \& Transformer) for the MIMIC-CXR dataset. In Report-to-Image retrieval, while MedViLL achieves lower performance than Bi and S2S with a statistically significant difference in MIMIC-CXR, MedViLL outperforms all baselines when fine-tuned on the unseen Open-I dataset except for Bi. However, although Bi outperforms MedViLL in Open-I, there is no statistically significant difference between both models with a p-value of 0.843. In Image-to-Report retrieval, S2S statistically outperforms MedViLL in MIMIC-CXR, but MedViLL is superior to all baselines with a statistically meaningful difference in Open-I. It is notable that the naive baselines, while severely underperforming for MIMIC-CXR, show substantially increased performance for Open-I. We believe this is due to the Open-I being a significantly smaller dataset than MIMIC-CXR, with only two Chexpert labels (Others, and Cardiomegaly) mostly dominating the label space.

\subsubsection{Medical Visual Question Answering}
{\hyperref[tab:vqa]{TABLE IV} shows the VQA accuracy when models were fine-tuned with all image types (‘ALL’), and with only the Chest X-ray images (‘CHEST’). We can see that MedViLL significantly outperforms MEVF \cite{nguyen2019overcoming}, the state-of-the-art model for the VQA-RAD dataset, indicating the effectiveness of the multi-modal pre-training for this complex multi-modal reasoning task. We can also see that MedViLL shows comparable performance as the bidirectional mask which allows unrestricted interaction between all text and vision features. In a statistical analysis, MedViLL shows significantly higher performance than all models for O.E. (open-ended questions) of 'ALL' and 'CHEST'. However, there was no statistically significant difference for Bi\&S2S of 'CHEST' and Bi of 'ALL', (p-values of 0.164, and 0.124, respectively). For C.E. (close-ended questions) of 'ALL',  Bi performed the best of all the models, followed by MedViLL. However, this result is not statistically meaningful, obtaining a p-value of 0.643 between Bi and MedViLL. Also, for C.E. of 'CHEST', MedViLL outperforms all baselines, but it is not statistically significant against Bi\&S2S (p-value of 0.519) and Bi (p-value of 0.549).}

\begin{figure*}[ht!]
\captionsetup{font=footnotesize}
\centering
\begin{subfigure}[ht!]{9cm}
\centering
\includegraphics[height=8cm, width=8cm]{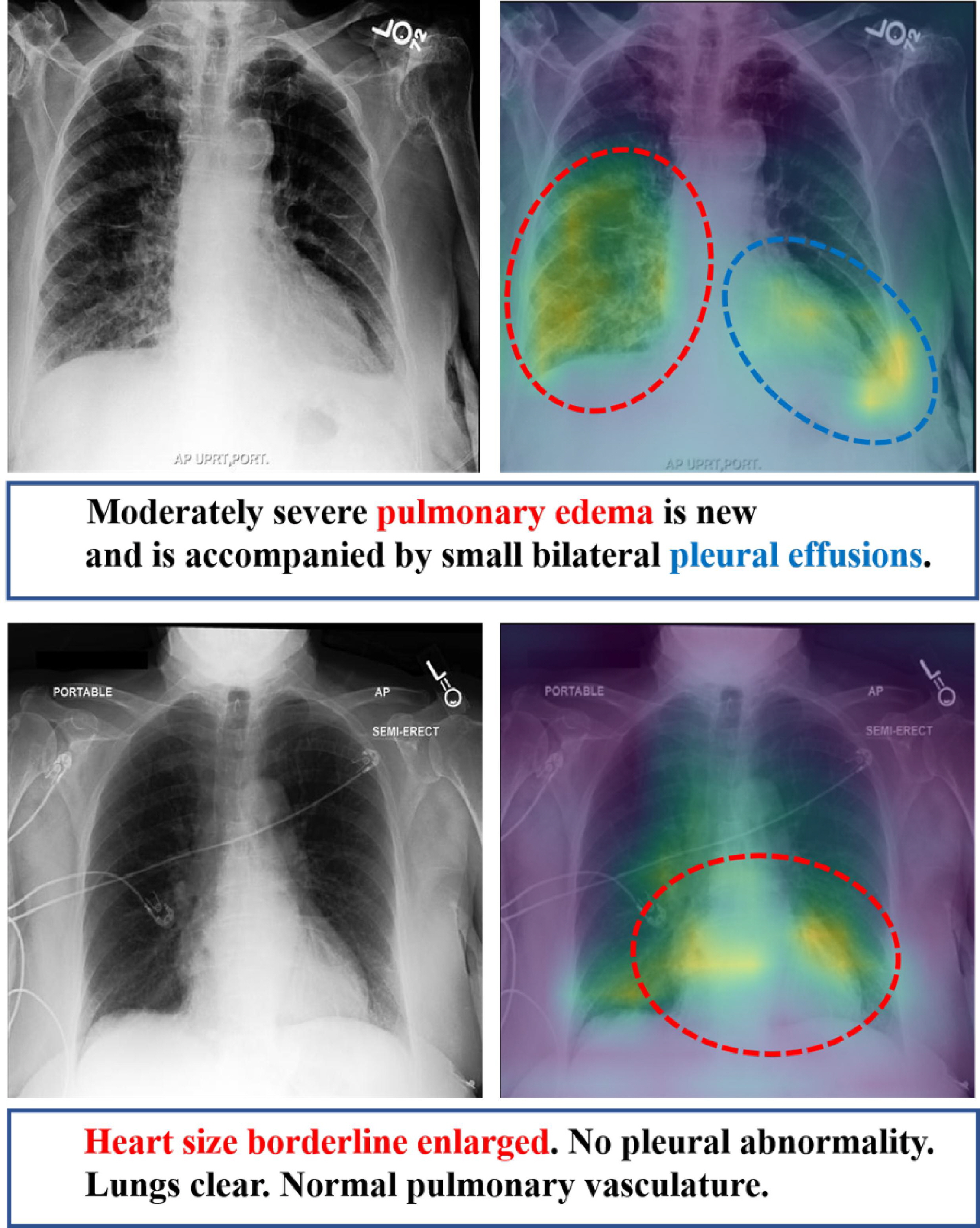}
\caption{Attention map visualization.}
\label{subfig:fig6}
\end{subfigure}
\begin{subfigure}[ht!]{9cm}
\centering
\includegraphics[height=8cm, width=8cm]{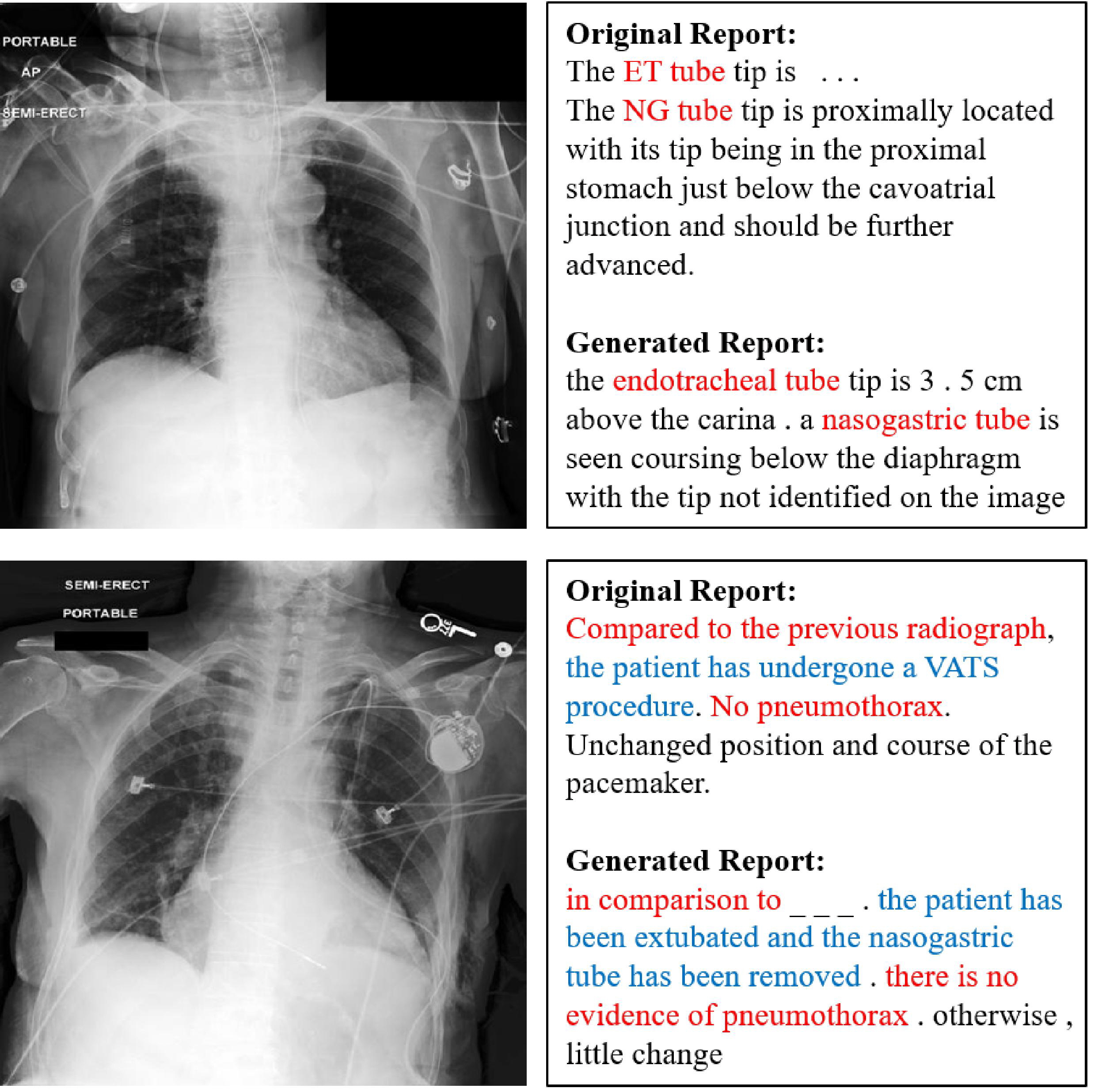}
\caption{Radiology report generation analysis.}
\label{subfig:report}
\end{subfigure}
\caption{\textbf{Qualitative results and analysis.} We visualize the attention regions extracted from the MedViLL (a). Also, we compare the generated report with the original report on the same chest X-ray image (b).}
\label{fig:fig7}
\end{figure*}

\begin{figure*}[ht!]
\captionsetup{font=footnotesize}
\centering
\includegraphics[height=6cm, width=\textwidth]{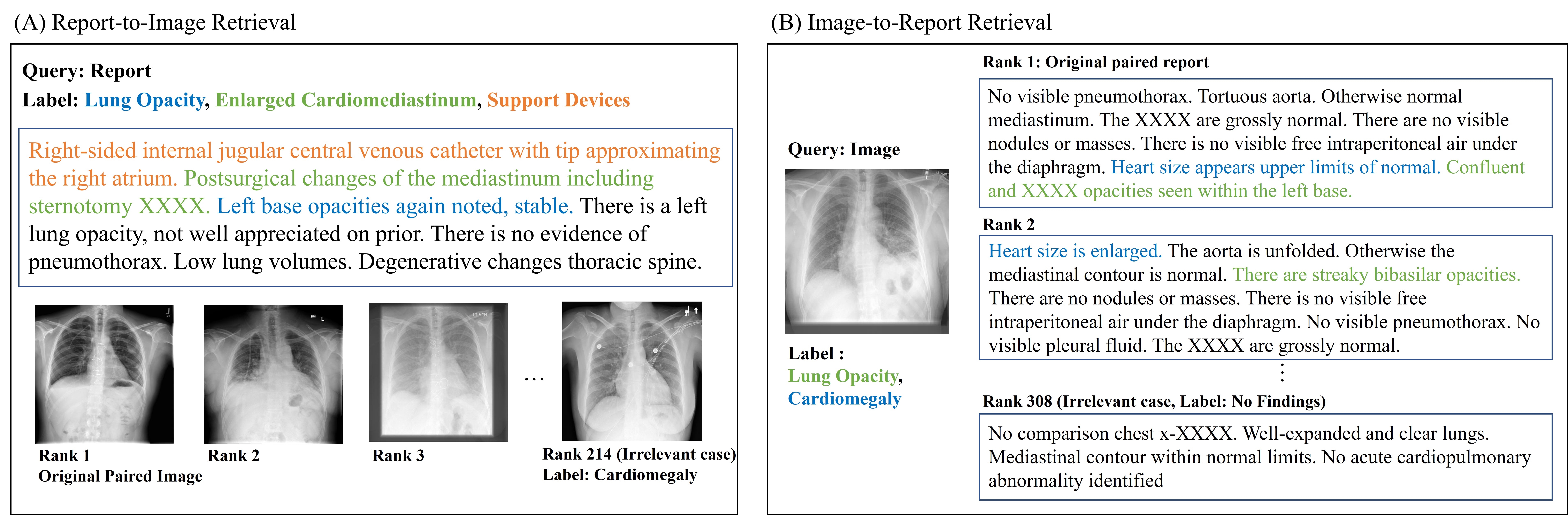}
\caption{\textbf{Case study of Medical Image-Report Retrieval.} Given a report or image as a query, we show the retrieved images or reports in (a) and (b) respectively. In (a), the given report is annotated by the Chexpert labeler with the diagnosis labels Lung Opacity (blue), Enlarged Cardiomediastinum (green), and Support Devices (orange). The top three images also contain the same labels as the given query while the last image (Rank 214) is irrelevant to the give query. In (B), the top three reports also contain the same labels recognized by the Chexpert labeler, Lung Opacity (green) and Cardiomegaly (blue).  Note that the last report (Rank 308) does not contain any label, and is irrelevant to the given query image.}
\label{fig:fig8}
\end{figure*}

\subsubsection{Radiology Report Generation}
\hyperref[tab:generation]{TABLE V} shows the report generation performance of all models. For this task, we also implemented TieNet \cite{wang2018tienet} as a baseline, which is a widely used CNN-RNN based attention model for report generation. We can see that the models pre-trained with auto-regressive manners (MedViLL, Bi\&S2S, S2S) all significantly outperform the other models in terms of both perplexity (except for TieNet) and clinical efficacy metrics for the MIMIC-CXR dataset. Among the clinical efficacy metrics, we report the p-value of F1 score because the F1 score is a balanced measure of both precision and recall and allows us to better capture the true performance here in light of the strong class imbalance of dataset. MedViLL achieved the best performance on the MIMIC dataset with a statistically significant difference. MedViLL seems to best capture the semantics embedded in the image given the highest F1 score. When fine-tuned on the unseen Open-I dataset, TieNet performed the best of all the models, followed by MedViLL. However, this result is not statistically significant between TieNet and MedViLL with a p-value of 0.2181, indicating that the performance of the two models is similar. As opposed to its generally favorable performance in VLU tasks, the bidirectional mask seems to be evidently harmful for the VLG task, most likely due to its incompatibility with the auto-regressive nature of VLG. Interestingly, we found N-gram based measures to be a suboptimal measure for report generation by showing that all models except TieNet achieved low BLEU scores; where the original report contains abbreviated terms (e.g. “ET tube”) models would generate expanded terms (e.g. “endotracheal tube”) and vice versa.

\noindent
\subsection{Qualitative results and analysis}
\subsubsection{Attention Map Visualization}
{We visualize the attention maps of the intermediate Transformer layers of MedViLL in order to gain qualitative insight of the cross-modality alignment between text tokens and image features as shown in \hyperref[fig:fig7]{Fig. 6} (a). Although MedViLL uses only report and images for training without any annotations, it can well attend to the disease-discriminatory regions written in the reports, as confirmed by a professional cardiologist. This suggests the potential of MedViLL’s capability to explain its learned representations to human users for smoother real-world deployment.}

\subsubsection{Radiology Report Generation}
{We compare the generated reports with the original report. As confirmed by a professional cardiologist, \hyperref[fig:fig7]{Fig. 6} (b) shows that MedViLL is able to generate clinically appropriate reports. Specifically, the left image-report pair of \hyperref[fig:fig7]{Fig. 6} (b) shows that MedViLL generates a report with the abbreviated medical terms (i.e. “ET tube”, “NG tube”) expanded to the original terms (i.e. “endotracheal tube” or “nasogastric tube”).  In the right image-report pair, the blue text in the original report describes the completion of VATS (i.e. video-assisted thoracic surgery). Interestingly, the generated report describes extubation and nasogastric tube removal, which is a part of VATS. This indicates that the BLEU score is not an appropriate measure to evaluate report generation especially in the medical domain.}

\subsubsection{Medical Image-Report Retrieval}
{We retrieved the cases pooled from 1,536 studies in the test set (\hyperref[fig:fig8]{Fig. 7}). As confirmed by a cardiologist, \hyperref[fig:fig8]{Fig. 7} demonstrate the clinical understanding of MedViLL. Specifically, we can observe that the results in the top-3 retrieved samples all share the same diagnosis labels as the given query; all top three images in \hyperref[fig:fig8]{Fig. 7} (a) are labeled with “Lung Opacity”, “Enlarged Cardiomediastinum” and “Support Devices” as the query report, and all top three reports in \hyperref[fig:fig8]{Fig. 7} (b) contain the same labels “Cardiomegaly” and “Lung Opacity” as the query image. Note that the samples in the low rank contain labels irrelevant to the given query.}

\section{Conclusion}
\label{sec:conclusion}
In this study, we propose a multi-modal pre-training model MedViLL, which uses a novel self-attention scheme to flexibly adapt to multiple downstream tasks of vision-language understanding and generation. By statistically and rigorously evaluating MedViLL on all four downstream tasks with three radiographic image-report datasets, we empirically demonstrated the superior performance of MedViLL against various baselines including task-specific architectures. Despite the impressive performance of MedViLL, this is just the beginning of the vision-language representation learning in the medical domain, and we plan to expand this approach to more diverse settings such as multi-view Chest X-ray studies or a sequence of studies over time.

\section*{Acknowledgment}
This work was supported by Samsung Electronics (No.IO201211-08109-01), Institute of Information \& Communications Technology Planning \& Evaluation (IITP) grant (No.2019-0-00075, Artificial Intelligence Graduate School Program(KAIST)), and National Research Foundation of Korea (NRF) grant (NRF-2020H1D3A2A03100945) funded by the Korea government (MSIT).

\bibliographystyle{abbrv}
\bibliography{reference_cite}

\newpage
\section{Appendix}
\subsection{Ablation studies}
\subsubsection{Visual Features}
To demonstrate the effectiveness of visual features, we performed additional experiments with visual feature sampling methods (full sample vs random 80\% sample) and various visual feature extractors (ResNet-50 vs ResNet-101 \cite{he2016deep} vs DenseNet-121 \cite{huang2017densely}). The detailed training method for each model is the same as that of MedViLL. We report the mean and standard deviation through 5 times random bootstrap experiments. As shown in \hyperref[tab:ablation_classificaiton]{TABLE VI}, \hyperref[tab:ablation_retrieval]{TABLE VII}, \hyperref[tab:ablation_vqa]{TABLE VIII}, and \hyperref[tab:ablation_generation]{TABLE IX}, all results reached poor performance in both in-domain and out-of-domain evaluation when using all the visual features of the CNN. These results show the difficulty of reaching good performance by over-fitting the MIMIC-CXR training set. Also, there are performance differences for each downstream task when using various visual feature extractor. We believe that better performance can be obtained with various experiments on the model architecture or hyper-parameter tuning (e.g., various visual feature extractors, BERT transforms, etc.)

% \subsubsection{Visual Backbone}
% To compare the visual feature extraction model, we further experimented with DenseNet-121 \cite{huang2017densely} and ResNet-101 \cite{he2016deep}. The detailed training method for each model is the same as that of MedViLL (e.g., 80\% random sample of visual features etc.), but only the CNN backbone was replaced for training and evaluation. We conduct 5 random bootstrap experiments and report the mean performance and its standard deviation.
% there is a performance difference for each downstream task. We believe that better performance can be obtained with various experiments on the model architecture or hyper-parameter tuning (e.g., various visual feature extractors, BERT transforms, etc.).

\subsubsection{Self-attention mask}
\begin{figure*}[htb!]
\captionsetup{font=footnotesize}
\centerline{\includegraphics[height=5.5cm, width=0.7\textwidth]{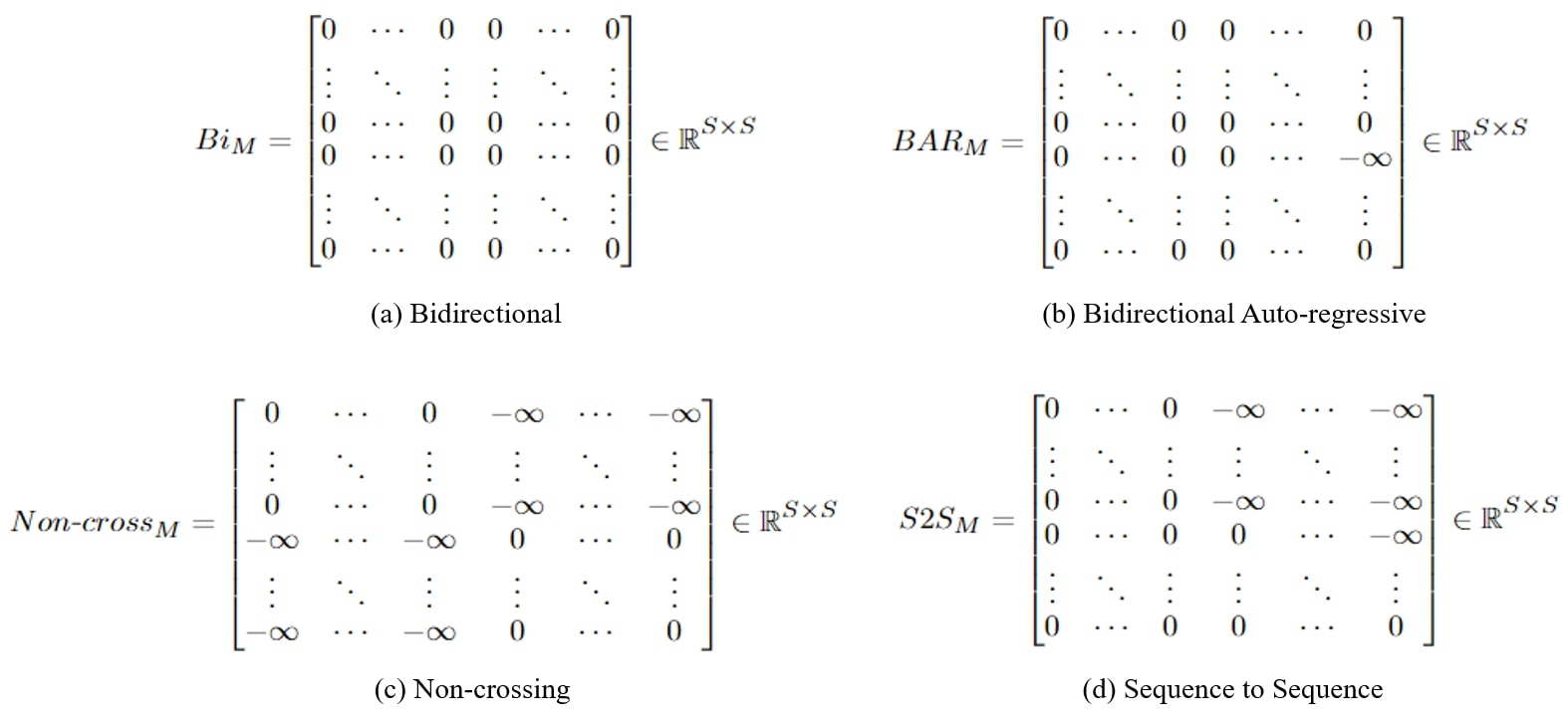}}
\caption{\textbf{Self-attention mask schemes.} Four types of the self-attention masks in matrix. The self-attention mask $M\in R^{S\times S}, S\,\text{=}\,visual\,features + language\,features + 3 special\,tokens$ consists of 0s and negative infinities.}
\label{fig:attn_matrix}
\end{figure*}

We explore the remaining 3 types of self-attention masks in this section. Bidirectional attention mask ((a) of \hyperref[fig:fig4]{Fig. 4} and \hyperref[fig:attn_matrix]{Fig. 8}) allows the attention calculation of all possible combinations (Eq. (10) - (13)) that encourage unconstrained context learning between vision and language modalities. This type of mask drives learning to understand the input correlation between the two vision-language modalities. On the other hand, Sequence-to-Sequence attention mask ((d) of \hyperref[fig:fig4]{Fig. 4} and \hyperref[fig:attn_matrix]{Fig. 8}) allows the attention calculation only to key from the visual modality (Eq. (10), (12)) and restricts all visual queries to attend any language key so that attention can be computed on input features sequentially. A model trained in this way can generate future features based on given input features in an auto-regressive manner. To restricts the interaction between two modalities, the Non-crossing attention mask ((c) of \hyperref[fig:fig4]{Fig. 4} and \hyperref[fig:attn_matrix]{Fig. 8}) only allows the attention calculation of each modality (Eq. (10), (12)).

\begin{table*}[ht]
\footnotesize
\captionsetup{font=footnotesize}
\centering
\caption{AUROC and F1 scores for the diagnosis classification task on MIMIC-CXR and Open-I. * indicates using all visual features.}
\setlength{\tabcolsep}{3pt}
% \vspace{-3mm}
\begin{tabular}{cccccc}
    \toprule
    Dataset & Metrics & \model & \model* & \model(Resnet101) & \model(Densenet121) \\\hline
    MIMIC-CXR & avg AUROC & \textbf{0.982 (0.00)} & 0.926 (0.0)& 0.918 (0.0)& 0.962 (0.0)\\
     & avg F1 & \textbf{0.841 (0.01)} & 0.699 (0.0) & 0.666 (0.01) & 0.771 (0.0)\\
    \midrule
    Open-I & avg AUROC	& 0.894 (0.00) & 0.886 (0.00)& 0.894 (0.00) & \textbf{0.897 (0.00)}\\
    & avg F1 & 0.408 (0.01) & 0.397 (0.00) & 0.404 (0.00) & \textbf{0.409 (0.00)}\\
    \bottomrule
\end{tabular}
\label{tab:ablation_classificaiton}
\end{table*}%
\noindent
\begin{table*}[h]
\footnotesize
\captionsetup{font=footnotesize}
\centering
\caption{Medical Image-Report Retrieval performance on MIMIC-CXR and Open-I. * indicates using all visual features.}
\setlength{\tabcolsep}{3pt}
% \vspace{-3mm}
\begin{tabular}{cccccccccc}
\toprule
    \multirow{2}{*}[-4pt]{Task} &
    \multirow{2}{*}[-4pt]{Models} &
    \multicolumn{4}{c}{MIMIC-CXR} &
    \multicolumn{4}{c}{OpenI} \\
    \cmidrule(lr){3-6}
    \cmidrule(lr){7-10}
    & & MRR & H@5 & R@5 & P@5 & MRR & H@5 & R@5 & P@5\\
    \midrule
    
    & \model & \textbf{58.1 (0.4)} &  \textbf{78.4 (0.7)} &  \textbf{47.8 (0.5)} &  \textbf{20.0 (0.1)} & \textbf{56.5 (1.3)} &  \textbf{74.8 (1.4)} & 13.3 (0.3) &  \textbf{37.0 (0.6)} \\
    
   Report-to-Image & \model* & 49.0 (0.1) & 69.7 (0.1) & 41.9 (0.1)& 17.6 (0.1)& 54.8 (0.1) &74.7 (0.1)& 13.4 (0.1)& 35.6 (0.1) \\
    
    & \model(Resnet101) &47.5 (0.1)  &  69.3 (0.1) & 41.5 (0.1) & 17.4 (0.1) & 53.1 (0.1) & 73.2 (0.1) & 12.8 (0.1) & 35.8 (0.1)\\
    
    & \model (Densenet121) & 49.3 (0.1) & 70.0 (0.1) & 42.2 (0.1) & 17.75 (0.1) & 56.4 (0.1)& 74.5 (0.1)& \textbf{14.0 (0.1)} & 36.7 (0.1)\\
    
    \midrule
    & \model & \textbf{56.1 (0.2)} & \textbf{76.2 (0.6)} & \textbf{48.1 (0.7)} & \textbf{19.9 (0.6)} & \textbf{49.3 (0.9)} &  \textbf{62.2 (0.7)} &  12.4 (0.2) &  34.2 (0.5)\\
    
    Image-to-Report & \model*  & 48.6 (0.1) & 68.0 (0.1) & 42.4 (0.1) & 17.4 (0.1) & 48.5 (0.1) & \textbf{62.2 (0.1)}& 12.5 (0.1)& \textbf{34.3 (0.1)}\\
    
    & \model(Resnet101)  & 46.3 (0.1) & 65.5 (0.1) & 39.8 (0.1) & 17.1 (0.1) & 47.2 (0.1) & 62.0 (0.1)  & 11.8 (0.1) & 32.5 (0.1)\\
    
    & \model (Densenet121) & 48.6 (0.1)  & 69.4 (0.1) & 42.7 (0.1) & 18.38 (0.1) &46.5 (0.1)& 60.9 (0.1)& \textbf{12.8 (0.1)} & 31.7 (0.1)\\
    
\bottomrule
\end{tabular}
\label{tab:ablation_retrieval}
\end{table*}%
\noindent
\begin{table*}[h]
\footnotesize
\captionsetup{font=footnotesize}
\centering
\caption{Model accuracy on the VQA-RAD dataset. * indicates using all visual features.}
\setlength{\tabcolsep}{3pt}
% \vspace{-3mm}

\begin{tabular}{cccccccccc}
\toprule
    \multirow{2}{*}[-4pt]{Models} &
    \multicolumn{2}{c}{ALL} &
    \multicolumn{2}{c}{CHEST} \\
    \cmidrule(lr){2-3}
    \cmidrule(lr){4-5}
             & Open-ended & Close-ended & Open-ended & Close-ended\\
    \midrule
    \model & \textbf{0.597(0.038)} & 0.782(0.022) & 0.608(0.051) & \textbf{0.783(0.033)} \\
    \model* & 0.512(0.012) & 0.743(0.009) & 0.572(0.021) & 0.736(0.006)\\
    \model(Resnet101) & 0.548(0.019)  & 0.781(0.002)  &  0.602(0.012) & 0.773(0.022) \\
    \model(Densenet121) & 0.593(0.001)  & \textbf{0.787(0.004)}  &  \textbf{0.612(0.01)} & 0.781(0.003) \\
\bottomrule
\end{tabular}
%}
\label{tab:ablation_vqa}
\end{table*}%

\noindent
\begin{table*}[htbp]
\footnotesize
\captionsetup{font=footnotesize}
\centering
\caption{Report generation performance in terms of Perplexity and Label Accuracy, Precision Recall and F1 and BLEU4. * indicates using all visual features.}
\setlength{\tabcolsep}{3pt}
% \vspace{-3mm}
\begin{tabular}{cccccccc}
\toprule
     Dataset & Models & Perplexity ($\downarrow$) & Accuracy ($\uparrow$) & Precision ($\uparrow$) & Recall ($\uparrow$) & F1 Score ($\uparrow$) & BLEU4 ($\uparrow$)\\
    \hline
    &\model & 4.19(0.03) & \textbf{0.841(0.003)} &  \textbf{0.698(0.002)} & \textbf{0.558(0.004)} & \textbf{0.620(0.010)} & 0.066(0.001)\\
    MIMIC &\model* & 5.04(0.01)& 0.780(0.016)  & 0.643(0.038) &0.417(0.026) & 0.505(0.027)& 0.072(0.009)  \\
    &\model(Resnet101) &\textbf{3.91(0.003)}&0.833(0.02)&0.652(0.004)&0.472(0.024)&0.526(0.003)&0.116(0.02)\\
    &\model(Densenet121) & 4.02(0.002)&0.81(0.003)&0.628(0.02)&0.425(0.004)&0.484(0.005)&\textbf{0.126(0.003)}\\

    \midrule
    &\model & 5.58(0.32) & 0.734(0.001) & 0.512(0.003) & \textbf{0.594(0.001)} & \textbf{0.550(0.009)} & 0.049(0.001)\\
    Open-I &\model* & 5.06(0.081)  & 0.790(0.018) & 0.577(0.015) & 0.294(0.089)& 0.382(0.081) & 0.044(0.010)  \\
    &\model(Resnet101) & \textbf{3.815(0.002)}  & 0.812(0.003) & \textbf{0.675(0.006)} & 0.367(0.003)& 0.466(0.001)& \textbf{0.075(0.031)} \\
    &\model(Densenet121) & 3.998(0.013) & \textbf{0.818(0.011)} & 0.582(0.03) &0.385(0.012) & 0.472(0.002) &  0.071(0.009) \\
\bottomrule
\end{tabular}
\label{tab:ablation_generation}
\end{table*}%

\noindent
\end{document}